\documentclass{article}

\usepackage[preprint]{neurips_2019}

\newcommand{\cf}{cf.~}

\newcommand{\eg}{e.g.~}

\newcommand{\ie}{i.e.~}

\usepackage{tikz}
\usetikzlibrary{positioning, shapes.geometric, backgrounds}
\definecolor{stateblue}{cmyk}{0.96,0,0,0}

\usepackage{subcaption}

\usepackage{grffile}

\usepackage{graphicx} %
\DeclareGraphicsExtensions{.pdf,.jpeg,.png}

\usepackage{rotating}

\usepackage{wrapfig}

\usepackage{amsmath}
\usepackage{amsthm}
\usepackage{amssymb}
\usepackage{amsfonts}       %
\usepackage{mathtools}
\usepackage{cancel}
\usepackage{bbm} %

\newcommand{\isd}{p_{s_0}} %

\newcommand{\adv}{\mathrm{a}} %

\newcommand{\B}[2]{\mathbb{B}^{#1} \left[ #2 \right]} %
\newcommand{\bigO}[1]{\mathcal{O}\left(#1\right)}
    \newcommand{\Ot}[1]{\tilde{\mathcal{O}}\left(#1\right)}

\newcommand{\diag}{\mathrm{diag}}

\newcommand{\E}[2]{\mathbb{E}_{#1} \left[ #2 \right]} %

\newcommand{\partdiff}[1]{\frac{\partial}{\partial #1}}

\newcommand{\piset}[1]{\Pi_{\mathrm{#1}}}
\newcommand{\prob}[1]{\mathrm{Pr}\{#1\}}

\newcommand{\real}[1]{\mathbb{R}^{#1}}
\newcommand{\integer}[1]{\mathbb{Z}^{#1}}

\newcommand{\setname}[1]{\mathcal{#1}}
\newcommand{\setsize}[1]{|\mathcal{#1}|}

\newcommand{\rmax}{r_{\mathrm{max}}}
\newcommand{\sref}{s_{\mathrm{ref}}}
\newcommand{\aref}{a_{\mathrm{ref}}}

\newcommand{\tmax}{t_{\mathrm{max}}}
\newcommand{\tmaxhat}{\hat{t}_{\mathrm{max}}} %

\newcommand{\gammabw}{\gamma_{\mathrm{Bw}}} %

\newcommand{\eqdef}{\coloneqq}
\newcommand{\eqdefr}{\eqqcolon}

\DeclareMathOperator*{\argmin}{\arg\!\min}
\DeclareMathOperator*{\argmax}{\arg\!\max}

\newcommand{\vecb}[1]{\boldsymbol{#1}}
\newcommand{\mat}[1]{\boldsymbol{#1}}

\let\originalleft\left
\let\originalright\right
\renewcommand{\left}{\mathopen{}\mathclose\bgroup\originalleft}
\renewcommand{\right}{\aftergroup\egroup\originalright}

\newcommand{\appref}[1]{Appendix~\ref{#1}} %

\newcommand{\eqreff}[1]{\eqref{#1}}
\newcommand{\eqreffand}[2]{(\ref{#1},~\ref{#2})}

\newcommand{\figref}[1]{Fig~\ref{#1}}

\newcommand{\figrefd}[4]{Figs~\ref{#1}, \ref{#2}, \ref{#3}, and \ref{#4}}

\newcommand{\secref}[1]{Sec~\ref{#1}}

\newcommand{\secrefand}[2]{Secs~\ref{#1} and \ref{#2}}

\newcommand{\tblref}[1]{Table~\ref{#1}}
\newcommand{\tblrefand}[2]{Tables~\ref{#1} and~\ref{#2}}

\newcommand{\alg}[1]{Algo~#1}

\newcommand{\app}[1]{Appendix~#1}
\newcommand{\assume}[1]{Assumption~#1}

\newcommand{\ch}[1]{Ch~#1}

\newcommand{\cor}[1]{Cor~#1} %

\newcommand{\equ}[1]{Eqn~#1}

\newcommand{\fig}[1]{Fig~#1}

\newcommand{\lmm}[1]{Lemma~#1}

\newcommand{\page}[1]{p#1}

\newcommand{\prop}[1]{Prop~#1}
\newcommand{\problem}[1]{Problem~#1}

\newcommand{\tbl}[1]{Table~#1}
\newcommand{\secc}[1]{Sec~#1}

\newcommand{\thm}[1]{Thm~#1}

\usepackage{booktabs}       %
\usepackage{multirow}
\usepackage{longtable}

\usepackage{enumitem}
\setlist{nosep}

\usepackage{enumitem}

\usepackage{xcolor}

\usepackage{pdflscape}
\usepackage{afterpage}

\usepackage{nicefrac}       %
\usepackage{microtype}      %
\usepackage[yyyymmdd,hhmmss]{datetime}
\usepackage{lipsum}

\usepackage[utf8]{inputenc} %
\usepackage[T1]{fontenc}    %

\usepackage{url}            %

\usepackage[hidelinks, bookmarksnumbered, backref=page]{hyperref}
\renewcommand*{\backref}[1]{}
\renewcommand*{\backrefalt}[4]{
  \ifcase #1 (Broken backref) %
  \or        (p#2)
  \else      (p#2)
  \fi
}
\hypersetup{
  colorlinks   = true,
  urlcolor     = blue,
  linkcolor    = blue,
  citecolor   = blue
}

\usepackage[linesnumbered,ruled,vlined]{algorithm2e}

\SetCommentSty{mycommfont}
\SetKwInput{KwInput}{Input}                %
\SetKwInput{KwOutput}{Output}              %

\usepackage{authblk}

\title{Average-reward model-free reinforcement learning: \\ a systematic review and literature mapping}

\author[1]{Vektor Dewanto}
\author[2]{George Dunn}
\author[2, 4]{Ali Eshragh}
\author[1]{Marcus Gallagher}
\author[3, 4]{Fred Roosta}
\affil[1]{School of Information Technology and Electrical Engineering, University of Queensland, AU}
\affil[2]{School of Mathematical and Physical Sciences, University of Newcastle, AU}
\affil[3]{School of Mathematics and Physics, University of Queensland, AU}
\affil[4]{International Computer Science Institute, Berkeley, CA, USA}
\affil[ ]{ \texttt{v.dewanto@uqconnect.edu.au, george.dunn@uon.edu.au,
  \{marcusg,fred.roosta\}@uq.edu.au, ali.eshragh@newcastle.edu.au} }

\begin{document}
\maketitle

\begin{abstract}
Reinforcement learning is important part of artificial intelligence.
In this paper, we review model-free reinforcement learning that
utilizes the average reward optimality criterion in the infinite horizon setting.
Motivated by the solo survey by \cite{mahadevan_1996_avgrew},
we provide an updated review of work in this area and
extend it to cover policy-iteration and function approximation methods
(in addition to the value-iteration and tabular counterparts).
We present a comprehensive literature mapping.
We also identify and discuss opportunities for future work.
\end{abstract}

\section{Introduction}

Reinforcement learning (RL) is one promising approach to
the problem of making sequential decisions under uncertainty.
Such a problem is often formulated as a Markov decision process (MDP)
with a state set~$\setname{S}$, an action set~$\setname{A}$,
a reward set~$\setname{R}$, and a decision-epoch set~$\setname{T}$.
At each decision-epoch (\ie timestep) $t \in \setname{T}$,
a decision maker (henceforth, an agent) is at a state $s_t \in \setname{S}$, and
chooses to then execute an action $a_t \in \setname{A}$.
Consequently, the agent arrives at the next state $s_{t+1}$ and earns an (immediate) reward $r_{t+1}$.
For $ t = 0, 1, \ldots, \tmax$ with $\tmax \le \infty$, the agent experiences a sequence of
$S_0, A_0, R_1, S_1, A_1, R_2, \ldots, S_{\tmax}$.
Here, $S_0$ is drawn from an initial state distribution~$\isd$, whereas
$s_{t+1}$ and $r_{t+1}$ are governed by the environment dynamics,
which is fully specified by
the state-transition probability $p(s' | s, a) \eqdef \prob{S_{t+1} = s' | S_t = s, A_t = a}$, and
the reward function $r(s, a, s') = \sum_{r \in \setname{R}} \prob{r | s, a, s'} \cdot r$.

The solution to the decision making problem is a mapping from every state to
a probability distribution over the set of available actions in that state.
This mapping is called a policy, \ie $\pi: \setname{S} \times \setname{A} \mapsto [0, 1]$,
where $\pi(a_t | s_t) \eqdef \prob{A = a_t | S_t = s_t}$.
Thus, solving such a problem amounts to finding the optimal policy, denoted by $\pi^*$.
The basic optimality criterion asserts that a policy with the largest value is optimal.
That is, $v(\pi^*) \ge v(\pi), \forall \pi \in \Pi$,
where the function $v$ measures the value of any policy $\pi$ in the policy set $\Pi$.
There are two major ways to value a policy based on the infinite reward sequence that it generates,
namely the average- and discounted-reward policy value formulations.
They induce the average- and discounted-reward optimality criteria, respectively.
For an examination of their relationship, pros, and cons in RL,
we refer the readers to \citep{dewanto_2021_avgdis}.

Furthermore, RL can be viewed as
simulation-based asynchronous approximate dynamic programming (DP)
\citep[\ch{7}]{gosavi_2015_sborl}.
Particularly in model-free RL, the simulation is deemed expensive because
it corresponds to direct interaction between an agent and its (real) environment.
Model-free RL mitigates not only the curse of dimensionality (inherently as approximate DP methods),
but also the curse of modeling (since no model learning is required).
This is in contrast to model-based RL, where an agent interacts with
the learned model of the environment \citep{dean_2017_lqr, jaksch_2010_ucrl2, tadepalli_1998_hlearning}.
Model-free RL is fundamental in that it encompasses the bare essentials for updating
sequential decision rules through natural agent-environment interaction.
The same update machinery can generally be applied to model-based RL.
In practice, we may always desire a system that runs both model-free and model-based mechanisms,
see \eg the so-called Dyna architecture \citep{silver_2008_dyna2, sutton_1990_dyna}.

This work surveys the existing value- and policy-iteration based
average-reward model-free RL.
We begin by reviewing relevant DP (as the root of RL), before progressing to
the tabular then function approximation settings in RL.
For comparison, the solo survey on average-reward RL \citep{mahadevan_1996_avgrew}
embraced only the tabular value-iteration based methods.
We present our review in \secrefand{sec:valiter}{sec:politer}, which are accompanied
by concise \tblrefand{tbl:valiter_work}{tbl:politer_work} (in \appref{sec:table_of_existing})
along with literature maps
(\figrefd{fig:taxonomy_optgainapprox}{fig:taxonomy_gain}{fig:taxonomy_stateval}{fig:taxonomy_actval}
in \appref{sec:taxonomy}).
We then discuss the insight and outlook in \secref{sec:discuss}.
Additionally in \appref{sec:problem_spec}, we compile environments
that were used for evaluation by the existing works.

In order to limit the scope, this work focuses on model-free RL
with a single non-hierarchical agent interacting \emph{online} with its environment.
We do not include works that \emph{approximately} optimize the average reward by
introducing a discount factor (hence, the corresponding approximation error), \eg
\cite{schneckenreither_2020_avgrew, karimi_2019_nasym, bartlett_2002_polgrad}.
We also choose not to examine RL methods that are based on
linear programming \citep{wang_2017_pilearn, neu_2017_entreg},
and decentralized learning automata \citep{chang_2009_dlfmc, wheeler_1986_dlmc}.

\section{Preliminaries} \label{sec:backgnd}

We assume that the problem we aim to tackle can be well modeled as
a Markov decision process (MDP) with the following properties.
\begin{itemize}
\item The state set $\setname{S}$ and action set $\setname{A}$ are finite.
All states are fully observable, and all actions $a \in \setname{A}$
are available in every state $s \in \setname{S}$.
The decision-epoch set $\setname{T} = \integer{\ge 0}$ is discrete and infinite.
Thus, we have discrete-time infinite-horizon finite MDPs.

\item The state-transition probability $p(s_{t+1}|s_t, a_t)$ and
the reward function $r(s_t, a_t)$ are both stationary (time homogenous, fixed over time).
Here, $r(s_t, a_t) = \E{}{r(s_t, a_t, S_{t+1})}$, and it is uniformly bounded by
a finite constant $\rmax$, \ie
$|r(s, a)| \le \rmax < \infty, \forall (s, a) \in \setname{S} \times \setname{A}$.

\item The MDP is recurrent, and
its optimal policies belong to the stationary policy set $\piset{S}$.
\end{itemize}

Furthermore, every stationary policy $\pi \in \piset{S}$ of an MDP induces
a Markov chain (MC), whose transition (stochastic) matrix is denoted as
$\mat{P}_\pi \in [0, 1]^{\setsize{S} \times \setsize{S}}$.
The $[s, s']$-entry of $\mat{P}_\pi$ indicates the probability of transitioning
from a current state $s$ to the next state $s'$ under a policy $\pi$.
That is, $\mat{P}_\pi[s, s'] = \sum_{a \in \setname{A}} \pi(a|s) p(s'|s, a)$.
The $t$-th power of $\mat{P}_\pi$ gives $\mat{P}_\pi^t$, whose $[s_0, s]$-entry
indicates the probability of being in state $s$ in $t$ timesteps
when starting from $s_0$ and following~$\pi$.
That is, $\mat{P}_\pi^t[s_0, s] = \prob{S_t = s | s_0, \pi} \eqdefr p_\pi^t(s | s_0)$.
The limiting distribution of $p_\pi^t$ is given by
\begin{equation}
p_\pi^\star(s | s_0)
= \lim_{\tmax \to \infty} \frac{1}{\tmax} \sum_{t=0}^{\tmax - 1} p_\pi^t(s | s_0)
= \lim_{\tmax \to \infty} p_\pi^{\tmax}(s | s_0),
\quad \forall s \in \setname{S},
\label{equ:pstar_lim}
\end{equation}
where the first limit is proven to exist in finite MDPs, while
the second limit exists whenever the finite MDP is aperiodic
\citep[\app{A.4}]{puterman_1994_mdp}.
This limiting state distribution $p_\pi^\star$ is equivalent to
the unique stationary (time-invariant) state distribution that satisfies
$(\vecb{p}_\pi^\star)^\intercal \mat{P}_\pi = (\vecb{p}_\pi^\star)^\intercal$,
which may be achieved in finite timesteps.
Here, $\vecb{p}_\pi^\star \in [0, 1]^{\setsize{S}}$ is
$p_\pi^\star(s | s_0)$ stacked together for all $s \in \setname{S}$.

The \emph{expected} average reward (also termed the gain) value of a policy $\pi$
is defined for all $s_0 \in \setname{S}$ as
\begin{align}
v_g(\pi, s_0)
& \eqdef \lim_{t_{\mathrm{max}} \to \infty} \frac{1}{t_{\mathrm{max}}}
    \E{S_t, A_t}{\sum_{t=0}^{t_{\mathrm{max}} - 1}
        r(S_t, A_t) \Big| S_0 = s_0, \pi}
    \label{equ:gain_lim} \\
& = \sum_{s \in \setname{S}}
    \Big\{ \lim_{\tmax \to \infty} \frac{1}{\tmax} \sum_{t=0}^{\tmax - 1}
        p_\pi^t(s | s_0) \Big\} r_\pi(s)
  = \sum_{s \in \setname{S}} p_\pi^\star(s| s_0) r_\pi(s)
    \label{equ:gain_prob} \\
& = \lim_{\tmax \to \infty} \E{S_{\tmax}, A_{\tmax}}{
        r(S_{\tmax}, A_{\tmax}) \Big| S_0 = s_0, \pi},
    \label{equ:gain_lim2}
\end{align}
where $r_\pi(s) = \sum_{a \in \setname{A}} \pi(a|s)\ r(s, a)$.
The limit in \eqref{equ:gain_lim} exists when the policy $\pi$ is stationary,
and the MDP is finite \citep[\prop{8.1.1}]{puterman_1994_mdp}.
Whenever it exists, the equality in \eqref{equ:gain_prob} follows due to
the existence of limit in \eqref{equ:pstar_lim} and
the validity of interchanging the limit and the expectation.
The equality in \eqref{equ:gain_lim2} holds if its limit exists, for instance
when $\pi$ is stationary and the induced MC is aperiodic
(nonetheless, note that even if the induced MC is periodic,
the limit in \eqref{equ:gain_lim2} exists for certain reward structures,
see \citet[\problem{5.2}]{puterman_1994_mdp}).
In matrix forms, the gain can be expressed as
\begin{equation*}
\vecb{v}_g(\pi)
= \lim_{\tmax \to \infty}  \frac{1}{\tmax} \vecb{v}_{\tmax}(\pi)
= \lim_{\tmax \to \infty}  \frac{1}{\tmax} \sum_{t=0}^{\tmax - 1} \mat{P}_\pi^t \vecb{r}_\pi
= \mat{P}_\pi^\star \vecb{r}_\pi,
\end{equation*}
where $\vecb{r}_\pi \in \real{\setsize{S}}$ is $r_\pi(s)$ stacked together for all $s \in \setname{S}$.
Notice that the gain involves taking the limit of the average of
the \emph{expected} total reward $\vecb{v}_{\tmax}$ from
$t=0$ to $\tmax - 1$ for $\tmax \to \infty$.

Since unichain MDPs have a single chain (\ie a closed irreducible recurrent class),
the stationary distribution is invariant to initial states.
Therefore, $\mat{P}_\pi^\star$ has identical rows so that the gain is
constant across all initial states.
That is, $v_g(\pi) = \vecb{p}_\pi^\star \cdot \vecb{r}_\pi = v_g(\pi, s_0), \forall s_0 \in \setname{S}$,
hence $\vecb{v}_g(\pi) = v_g(\pi) \cdot \vecb{1}$.
The gain $v_g(\pi)$ can be interpreted as the stationary reward because it
represents the average reward per timestep of a system in its steady-state under~$\pi$.

A policy $\pi^* \in \piset{S}$ is gain-optimal (hereafter, simply called optimal) if
\begin{equation}
v_g(\pi^*, \cancel{s_0}) \ge v_g(\pi, \cancel{s_0}),
\quad \forall \pi \in \piset{S}, \forall s_0 \in \setname{S},
\quad \text{hence}\ \pi^* \in \argmax_{\pi \in \piset{S}} v_g(\pi)
\label{equ:gainopt}
\end{equation}
Such an optimal policy $\pi^*$ is also a greedy (hence, deterministic) policy
in that it selects actions maximizing the RHS of
the average-reward Bellman \emph{optimality} equation
(which is useful for deriving optimal control algorithms) as follows,
\begin{equation}
v_b(\pi^*, s) + v_g(\pi^*) = \max_{a \in \setname{A}} \Big\{ r(s, a)
    + \sum_{s' \in \setname{S}} p(s' | s, a)\ v_b(\pi^*, s') \Big\},
\quad \forall s \in \setname{S},
\label{equ:avgrew_bellman_optim}
\end{equation}
where $v_b$ denotes the bias value.
It is defined for all $\pi \in \piset{S}$ and  all $s_0 \in \setname{S}$ as
\begin{align}
v_b(\pi, s_0)
& \eqdef \lim_{\tmax \to \infty} \E{S_t, A_t}{
    \sum_{t=0}^{\tmax - 1}
    \Big( r(S_t, A_t) - v_g(\pi) \Big) \Big| S_0 = s_0, \pi}
    \label{equ:bias_lim} \\
& = \lim_{\tmax \to \infty} \sum_{t=0}^{\tmax - 1}
    \sum_{s \in \setname{S}} \Big( p_\pi^t(s|s_0) - p_\pi^\star(s) \Big)  r_\pi(s)
    \label{equ:bias_prob2} \\
& = \underbrace{\sum_{t=0}^{\tau - 1}
        \sum_{s \in \setname{S}} p_\pi^t(s|s_0) r_\pi(s)
        }_\text{the expected total reward $v_\tau$}
    -\ \tau v_g(\pi)
    + \underbrace{\lim_{\tmax \to \infty} \sum_{t=\tau}^{\tmax - 1} \sum_{s \in \setname{S}}
        \Big( p_\pi^t(s|s_0) - p_\pi^\star(s) \Big)  r_\pi(s)
        }_\text{approaches 0 as $\tau \to \infty$}
    \label{equ:bias_prob3} \\
& = \sum_{s \in \setname{S}} \Big\{
    \lim_{\tmax \to \infty} \sum_{t=0}^{\tmax - 1}
    \Big( p_\pi^t(s|s_0) - p_\pi^\star(s) \Big) \Big\} r_\pi(s)
    = \sum_{s \in \setname{S}} d_\pi(s|s_0) r_\pi(s),
    \label{equ:bias_prob}
\end{align}
where all limits are assumed to exist, and
\eqref{equ:bias_lim} is bounded because of the subtraction of $v_g(\pi)$.
Whenever exchanging the limit and the expectation is valid in \eqref{equ:bias_prob},
$d_\pi(s|s_0)$ represents the $[s_0, s]$-entry of
the non-stochastic deviation matrix
$\mat{D}_\pi
\eqdef (\mat{I} - \mat{P}_\pi + \mat{P}_\pi^\star)^{-1} (\mat{I} - \mat{P}_\pi^\star)$.

The bias $v_b(\pi, s_0)$ can be interpreted in several ways.
\textbf{Firstly} based on \eqref{equ:bias_lim}, the bias is the expected total difference
between the immediate reward $r(s_t, a_t)$ and the stationary reward $v_g(\pi)$
when a process starts at~$s_0$ and follows~$\pi$.
\textbf{Secondly} from \eqref{equ:bias_prob2}, the bias indicates the difference between
the expected total rewards of two processes under $\pi$: one starts at $s_0$ and
the other at an initial state drawn from $p_\pi^\star$.
Put in another way, it is the difference of the total reward of $\pi$ and
the total reward that would be earned if the reward per timestep were $v_g(\pi)$.
\textbf{Thirdly}, decomposing the timesteps as in \eqref{equ:bias_prob3} yields
$v_\tau(\pi, s_0) \approx v_g(\pi) \tau + v_b(\pi, s_0)$.
This suggests that the bias serves as the intercept of a line around which
the expected total reward $v_\tau$ oscillates, and eventually converges as $\tau$ increases.
Such a line has a slope of the gain value.
For example in an MDP with zero reward absorbing terminal state (whose gain is 0),
the bias equals the expected total reward before the process is absorbed.
\textbf{Lastly}, the deviation factor $(p_\pi^t(s|s_0) - p_\pi^\star(s))$ in \eqref{equ:bias_prob}
is non-zero only before the process reaches its steady-state.
Therefore, the bias indicates the transient performance.
It may be regarded as the ``transient'' reward, whose
values are earned during the transient phase.

For any reference states $\sref \in \setname{S}$, we can define
the (bias) relative value $v_{brel}$ as follows
\begin{equation*}
v_{brel}(\pi, s)
\eqdef v_b(\pi, s) - v_b(\pi, \sref)
=  \lim_{\tmax \to \infty} \{ v_{\tmax}(\pi, s)  - v_{\tmax}(\pi, \sref) \},
\quad \forall \pi \in \piset{S}, \forall s\in \setname{S}.
\end{equation*}
The right-most equality follows from \eqref{equ:bias_lim} or \eqref{equ:bias_prob3},
since the gain are the same from both $s$ and $\sref$.
It indicates that $v_{brel}$ represents the asymptotic different
in the expected total reward due to starting from $s$, instead of $\sref$.
More importantly, the relative value $v_{brel}$ is equal to $v_b$
up to some constant for any $s \in \setname{S}$.
Moreover, $v_{brel}(\pi, s=\sref) = 0$.
After fixing $\sref$, therefore, we can substitute $v_{brel}$ for $v_b$ in
\eqref{equ:avgrew_bellman_optim}, then uniquely determine $v_{brel}$ for all states.
Note that \eqref{equ:avgrew_bellman_optim} with $v_b$ is originally
an underdetermined nonlinear system with $\setsize{S}$ equations and
$\setsize{S} + 1$ unknowns (\ie one extra of $v_g$).

In practice, we often resort to $v_{brel}$ and abuse the notation $v_b$ to
also denote the relative value whenever the context is clear.
In a similar fashion, we call the bias as the relative/differential value.
One may also refer to the bias as relative values due to \eqref{equ:bias_prob2},
average-adjusted values (due to \eqref{equ:bias_lim}), or
potential values (similar to potential energy in physics
whose values differ by a constant \citep[\page{193}]{cao_2007_slo}).

For brevity, we often use the following notations,
$v_g^\pi \eqdef v_g(\pi) \eqdefr g^\pi$, and $v_g^* \eqdef v_g(\pi^*) \eqdefr g^*$, as well as
$v_b^\pi(s) \eqdef v_b(\pi, s)$, and $v_b^*(s) \eqdef v_b(\pi^*, s)$.
Here, $v_b^\pi(s)$ may be read as the relative state value of $s$ under a policy~$\pi$.

\section{Value-iteration schemes} \label{sec:valiter}

Based on the Bellman optimality equation \eqref{equ:avgrew_bellman_optim},
we can obtain the optimal policy once we know the optimal state value
$\vecb{v}_b^* \in \real{\setsize{S}}$, which includes knowing the optimal gain $v_g^*$.
Therefore, one approach to optimal control is to (approximately) compute $\vecb{v}_b^*$,
which leads to the value-iteration algorithm in DP.
The value-iteration scheme in RL uses the same principle as its DP counterpart.
However, we approximate the optimal (state-)action value
$\vecb{q}_b^* \in \real{\setsize{S} \setsize{A}}$, instead of $\vecb{v}_b^*$.

In this section, we begin by introducing the foundations of the value-iteration scheme,
showing the progression from DP into RL.
We then cover tabular methods by presenting how to estimate the optimal action values iteratively
along with numerous approaches to estimating the optimal gain.
Lastly, we examine function approximation methods, where
the action value function is parameterized by a weight vector, and
present different techniques of updating the weights.
\tblref{tbl:valiter_work} (in \appref{sec:table_of_existing}) summarizes existing
average reward model-free RL based on the value-iteration scheme.

\subsection{Foundations} \label{sec:valiter_backgnd}

In DP, the basic value iteration (VI) algorithm \emph{iteratively} solves for
the optimal state value $\vecb{v}_b^*$ and outputs an $\varepsilon$-optimal policy;
hence it is deemed exact up to the convergence tolerance $\varepsilon$.
The basic VI algorithm proceeds with the following steps.
\begin{enumerate} [label=\textbf{Step~\arabic{*}:}]
\item Initialize the iteration index $k \gets 0$ and $\hat{v}_b^{k=0}(s) \gets 0$,
    where $\hat{v}_b^k(s) \approx v_b^*(s), \forall s \in \setname{S}$. \\
    Also select a small positive convergence tolerance $\varepsilon > 0$. \\
    Note that $\hat{v}_b^k$ may not correspond to any stationary policy.

\item Perform updates on \emph{all} iterates (termed as \emph{synchronous} updates)
    as follows,
    \begin{equation*}
    \hat{v}_b^{k+1}(s) = \B{*}{\hat{v}_b^{k}(s)}, \quad \forall s \in \setname{S},
    \tag{Basic VI: synchronous updates}
    \end{equation*}
    where the average-reward Bellman optimality operator $\mathbb{B}^*$ is defined as
    \begin{equation*}
        \B{*}{\hat{v}_b^k(s)} \eqdef \max_{a \in \setname{A}}
            \Big[ r(s, a) + \sum_{s' \in \setname{S}} p(s' | s, a)\ \hat{v}_b^k(s') \Big],
        \quad \forall s \in \setname{S}.
    \end{equation*}
    Clearly, $\mathbb{B}^*$ is non-linear (due to the $\max$ operator), and
    is derived from \eqref{equ:avgrew_bellman_optim} with $\hat{v}_g^* \gets 0$.

\item If the span seminorm $sp(\hat{\vecb{v}}_b^{k+1} - \hat{\vecb{v}}_b^{k}) > \varepsilon$,
    then increment $k$ and go to Step~2. \\
    Otherwise, output a greedy policy with respect to the RHS of \eqref{equ:avgrew_bellman_optim};
    such a policy is $\varepsilon$-optimal.
    Here, $sp(\vecb{v}) \eqdef \max_{s \in \setname{S}} v(s) - \min_{s' \in \setname{S}} v(s')$,
    which measure the range of all components of a vector~$\vecb{v}$.
    Therefore, although the vector changes, its span may be constant.
\end{enumerate}

The mapping in Step~2 of the VI algorithm is not contractive (because there is no discount factor).
Moreover, the iterates $\hat{\vecb{v}}_b^k$ can grow very large, leading to numerical instability.
Therefore, \cite{white_1963_dpmc} proposed substracting out the value of
an arbitrary-but-fixed reference state~$\sref$ at every iteration.
That is,
\begin{equation*}
\hat{v}_b^{k+1}(s) = \B{*}{\hat{v}_b^{k}(s)} - \hat{v}_b^{k}(\sref),
\quad \forall s \in \setname{S},
\tag{Relative VI: synchronous updates}
\end{equation*}
which results in the so-called \emph{relative} value iteration (RVI) algorithm.
This update yields the same span and the same sequence of maximizing actions as
that of the basic VI algorithm.
Importantly, as $k \to \infty$, the iterate $\hat{v}_b^{k}(\sref)$ converges to $v_g^*$.

The \emph{asynchronous} version of RVI may diverge
\citetext{\citealp[\page{682}]{abounadi_2001_rviqlearn}, \citealp[\page{232}]{gosavi_2015_sborl}}.
As a remedy, \cite{jalali_1990_avgcost} introduced the following update,
\begin{equation*}
\hat{v}_b^{k+1}(s) = \B{*}{\hat{v}_b^{k}(s)} - \hat{v}_g^k,
\quad \forall (s \ne \sref) \in \setname{S},
\tag{Relative VI: asynchronous updates}
\end{equation*}
where $\hat{v}_g^k$ is the estimate of $v_g^*$ at iteration~$k$, and
$\hat{v}_b^{k}(\sref) = 0$, for all iterations $k = 0, 1, \ldots$.
It is shown that this asynchronous method converges to produce a gain optimal policy
in a regenerative process, where there exists a \emph{single} recurrent state
under \emph{all} stationary and deterministic policies.

In RL, the reward function $r(s, a)$ and the transition distribution $p(s'|s, a)$ are unknown.
Therefore, it is convenient to define a relative (state-)action value $\vecb{q}_b(\pi)$
of a policy $\pi$ as follows,
\begin{align}
q_b(\pi, s, a)
& \eqdef \lim_{\tmax \to \infty} \E{S_t, A_t}{\sum_{t=0}^{\tmax - 1} \left( r(S_t, A_t)
    - v_g^\pi \right) \Big| S_0 = s, A_0 = a, \pi} \notag \\
& = r(s, a) - v_g^\pi + \E{}{v_b^\pi(S_{t+1}) },
\qquad \forall (s, a) \in \setname{S} \times \setname{A},
\label{equ:qb}
\end{align}
where for brevity, we define $q_b^\pi(s, a) \eqdef q_b(\pi, s, a)$,
as well as $q_b^*(s, a) \eqdef q_b(\pi^*, s, a)$.
The corresponding average reward Bellman optimality equation is
as follows \citep[\page{549}]{bertsekas_2012_dpoc},
\begin{equation}
q_b^*(s, a) + v_g^* =
    r(s, a)
    + \sum_{s' \in \setname{S}} p(s' | s, a)
        \underbrace{\max_{a' \in \setname{A}} q_b^*(s', a'),}_\text{$v_b^*(s')$}
\quad \forall (s, a) \in \setname{S} \times \setname{A},
\label{equ:avgrew_bellmanopt_q}
\end{equation}
whose RHS is in the same form as the quantity maximized over all actions
in \eqref{equ:avgrew_bellman_optim}.
Therefore, the gain-optimal policy $\pi^*$ can be obtained simply by acting greedily over
the optimal action value; hence, $\pi^*$ is deterministic.
That is,
\begin{equation*}
\pi^*(s) = \argmax_{a \in \setname{A}} \Big[
\underbrace{
    r(s, a)
    + \sum_{s' \in \setname{S}} p(s' | s, a) \max_{a' \in \setname{A}} q_b^*(s', a')
    - v_g^*
}_{q_b^*(s, a)}
\Big], \qquad \forall s \in \setname{S}.
\end{equation*}
As can be observed, $\vecb{q}_b^*$ combines the effect of $p(s'|s, a)$ and $\vecb{v}_b^*$,
without estimating them separately at the cost of an increased number of
estimated values since typically $\setsize{S} \times \setsize{A} \gg \setsize{S}$.
The benefit is that action selection via $\vecb{q}_b^*$ does not require
the knowledge of $r(s, a)$ and $p(s'|s, a)$.
Note that $v_g^*$ is invariant to state~$s$ and action~$a$;
hence its involvement in the above maximization has no effect.

Applying the idea of RVI on action values $\vecb{q}_b^*$ yields the following iterate,
\begin{align}
\hat{q}_b^{k+1}(s, a)
& = \B{*}{\hat{q}_b^{k}(s, a)} - v_b^{k}(\sref)
    \tag{Relative VI on $\vecb{q}_b^*$: asynchronous updates} \\
& = \underbrace{
        r(s, a) + \sum_{s' \in \setname{S}} p(s' | s, a)
        \max_{a' \in \setname{A}}  \hat{q}_b^k(s', a')
    }_{\B{*}{\hat{q}_b^{k}(s, a)}}
    - \underbrace{
        \max_{a'' \in \setname{A}} \hat{q}_b^k(\sref, a'')
        }_\text{can be interpreted as $\hat{v}_g^*$},
\label{equ:rviq_iter}
\end{align}
where $\hat{\vecb{q}}_b^{k}$ denotes the estimate of $\vecb{q}_b^*$ at iteration~$k$,
and $\mathbb{B}^*$ is based on \eqref{equ:avgrew_bellmanopt_q} so it operates on action values.
The iterates of $\hat{v}_b^{k}(s) = \max_{a \in \setname{A}} \hat{q}_b^k(s, a)$
are conjectured to converge to $v_b^*(s)$ for all $s \in \setname{S}$
by \citet[\secc{7.2.3}]{bertsekas_2012_dpoc}.

\subsection{Tabular methods} \label{sec:valiter_tabular}

In model-free RL, the iteration for estimating $\vecb{q}_b^*$ in \eqreff{equ:rviq_iter}
is carried out asynchronously as follows,
\begin{equation}
\hat{q}_b^*(s, a)
\gets \hat{q}_b^*(s, a)
    + \beta \big\{
        r(s, a) + \max_{a' \in \setname{A}} \hat{q}_b^*(s', a')
        - \hat{v}_g^* -\ \hat{q}_b^*(s, a)
    \big\},
\label{equ:qbstar_update}
\end{equation}
where $\beta$ is a positive stepsize, whereas $s$, $a$, and $s'$ denote
the current state, current action, and next state, respectively.
Here, the sum over $s'$ in \eqreff{equ:rviq_iter},
\ie the expectation with respect to $S'$, is approximated by a single sample $s'$.
The stochastic approximation (SA) based update in \eqreff{equ:qbstar_update} is
the essense of $Q_b$-learning \citep[\secc{5}]{schwartz_1993_rlearn}, and most of its variants.
One exception is that of \citet[\equ{8}]{prashanth_2011_tlrl},
where there is no subtraction of $\hat{q}_b^*(s, a)$.

In order to prevent the iterate $\hat{q}_b^*$ from becoming very large (causing numerical instability),
\citet[\page{702}]{singh_1994_avgrewrl} advocated assigning $q_b^*(\sref, \aref) \gets 0$,
for arbitrary-but-fixed reference state $\sref$ and action $\aref$.
Alternatively, \citet[\page{404}]{bertsekas_1996_neurodp} advised setting $\hat{q}_b^*(\sref, \cdot) \gets 0$.
Both suggestions seem to follow the heuristics of obtaining the unique solution of
the underdetermined Bellman optimality non-linear system of equations in \eqreff{equ:avgrew_bellman_optim}.

There are several ways to approximate the optimal gain $v_g^*$ in \eqreff{equ:qbstar_update},
as summarized in \figref{fig:taxonomy_optgainapprox} (\appref{sec:taxonomy}).
In particular, \citet[\secc{2.2}]{abounadi_2001_rviqlearn} proposed three variants as follows.
\begin{enumerate} [label=\roman{*}.]
\item $\hat{v}_g^* \gets \hat{q}_b^*(\sref, \aref)$
    with reference state $\sref$ and action $\aref$.
    \cite{yang_2016_csv} argued that properly choosing $\sref$ can be difficult
    in that the choice of $\sref$ affects the learning performance,
    especially when the state set is large.
    They proposed setting $\hat{v}_g^* \gets c$ for a constant $c$ from prior knowledge.
    Moreover, \cite{wan_2020_avgrew} showed empirically that such a reference
    retards learning and causes divergence.
    This happens when $(\sref, \aref)$ is infrequently visited,
    \eg being a trasient state in unichain MDPs.

\item $\hat{v}_g^* \gets \max_{a' \in \setname{A}} \hat{q}_b^*(s_\mathrm{ref}, a')$,
    used by \citet[\equ{8}]{prashanth_2011_tlrl}.
    This inherits the same issue regarding $\sref$ as before.
    Whenever the action set is large, the maximization over $\setname{A}$ should
    be estimated, yielding another layer of approximation errors.

\item $\hat{v}_g^* \gets
    \sum_{(s, a) \in \setname{S} \times \setname{A}} \hat{q}_b^*(s, a) / (\setsize{S} \setsize{A})$,
    used by \citet[\equ{11}]{avrachenkov_2020_wiql}.
    Averaging all entries of $\hat{\vecb{q}}_b^*$ removes the need for $\sref$ and $\aref$.
    However, because $\hat{q}_b^*$ itself is an estimating function with diverse accuracy
    across all state-action pairs, the estimate $\hat{v}_g^*$ involves the averaged approximation error.
    The potential issue due to large state and action sets is also concerning.
\end{enumerate}
Equation \eqreff{equ:qbstar_update} with one of three proposals (i - iii) for $v_g^*$ estimators
constitutes the RVI $Q_b$-learning.
Although it operates asynchronously, its convergence is assured by decreasing the stepsize~$\beta$.

The optimal gain $v_g^*$ in \eqreff{equ:qbstar_update} can also be estimated
iteratively via SA as follows,
\begin{equation}
\hat{v}_g^* \gets \hat{v}_g^* + \beta_g \Delta_g,
\quad \text{for some update $\Delta_g$ and a positive stepsize $\beta_g$}.
\label{equ:vgstar_sa}
\end{equation}
Thus, the corresponding $Q_b$-learning becomes 2-timescale SA,
involving both $\beta$ and $\beta_g$.
There exist several variations on $\Delta_g$ as listed below.
\begin{enumerate} [label=\roman{*}.]
\item In \citet[\secc{5}]{schwartz_1993_rlearn},
    \begin{equation}
    \Delta_g \eqdef r(s, a) +
        \underbrace{
            \max_{a' \in \setname{A}} \hat{q}_b^*(s', a')
            - \max_{a' \in \setname{A}} \hat{q}_b^*(s, a')
        }_\text{to minimize the variance of updates}
        -\ \hat{v}_g^*,
        \quad \text{if}\
        \underbrace{a = \argmax_{a' \in \setname{A}} \hat{q}_b^*(s, a')}_\text{a greedy action $a$}.
        \label{equ:vgstarhat_schwartz}
    \end{equation}
    By updating only when greedy actions are executed, the influence of exploratory actions
    (which are mostly suboptimal) can be avoided.

\item In \citet[\alg{3}]{singh_1994_avgrewrl}, and \citet[\equ{6}]{wan_2020_avgrew},
    \begin{equation*}
    \Delta_g \eqdef \underbrace{
            r(s, a)
            + \max_{a' \in \setname{A}} \hat{q}_b^*(s', a')
            - \hat{q}_b^*(s, a)
        }_\text{$v_g^*$ in expectation of $S'$ when using $q_b^*$
            (see \eqreff{equ:avgrew_bellmanopt_q})}
        -\ \hat{v}_g^*. %
        \tag{To update at every action}
    \end{equation*}
    Since the equation for $v_g^*$ \eqreff{equ:avgrew_bellmanopt_q} applies to any state-action pairs,
    it is reasonable to update $\hat{v}_g^*$ for both greedy and exploratory actions as above.
    This also implies that information from non-greedy actions is not wasted.
    Hence, it is more sample-efficient than that of \eqreff{equ:vgstarhat_schwartz}.

\item In \citet[\alg{4}]{singh_1994_avgrewrl}, and \citet[\equ{8}]{das_1999_smart},
    \begin{equation*}
        \Delta_g \eqdef r(s, a) - \hat{v}_g^*,
            \quad \text{if $a$ is chosen greedily with $\beta_g$ set to $1/(n_u + 1)$},
    \end{equation*}
    where $n_u$ denotes the number of $\hat{v}_g^*$ updates so far \eqreff{equ:vgstar_sa}.
    This special value of $\beta_g$ makes the estimation equivalent to
    the sample average of the rewards received for greedy actions.

\item In \citet[\page{404}]{bertsekas_1996_neurodp}, \citet[\equ{2.9b}]{abounadi_2001_rviqlearn},
    and \citet[\page{551}]{bertsekas_2012_dpoc},
    \begin{equation}
        \Delta_g \eqdef \underbrace{
            \max_{a' \in \setname{A}} \hat{q}_b^*(\sref, a')
            }_{\hat{v}_b^*(\sref)},
        \quad \text{for an arbitrary reference state $\sref$}.
        \label{equ:optgain_sa_bertsekas}
    \end{equation}
    This benefits from having $\sref$ such that $v_b^*(\sref)$ can be interpreted as $v_g^*$,
    while also satisfying the underdetermined system of average-reward Bellman optimality equations.
\end{enumerate}

\subsection{Methods with function approximation}

We focus on \emph{parametric} techniques for $Q_b$-learning with function approximation.
In such cases, the action value is parameterized by
a weight (parameter) vector $\vecb{w} \in \setname{W} = \real{\dim(\vecb{w})}$,
where $\dim(\vecb{w})$ is the number of dimensions of $\vecb{w}$.
That is,
$\hat{q}_b^*(s, a; \vecb{w}) \approx q_b^*(s, a), \forall (s, a) \in \setname{S} \times \setname{A}$.
Note that there also exist non-parametric techniques, for instance,
kernel-based methods \citep{ormoneit_2002_kbrlavg, ormoneit_2001_kbrl}
and those based on state aggregation \citep{ortner_2007_pmet}.

\citet[\page{404}]{bertsekas_1996_neurodp} proposed the following weight update,
\begin{equation}
\vecb{w} \gets \vecb{w}
    + \beta \Big\{
        \Big( r(s, a) - \hat{v}_g^*
            +  \max_{a' \in \setname{A}} \hat{q}_b^*(s', a'; \vecb{w})
        \Big) \nabla \hat{q}_b^*(s, a; \vecb{w})
        - \vecb{w}
        \Big\}.
\label{equ:qbw_update_bertsekas}
\end{equation}
Particularly, the optimal gain $v_g^*$ is estimated using
\eqreffand{equ:vgstar_sa}{equ:optgain_sa_bertsekas}.
They highlighted that even if $\hat{q}_b^*(\vecb{w})$ is bounded,
$\hat{v}_g^*$ may diverge.

\citet[\equ{9}]{das_1999_smart} updated the weight using temporal difference
(TD, or TD error) as follows,
\begin{equation}
\vecb{w} \gets \vecb{w} + \beta \Big\{
    \Big(
    \underbrace{
        r(s, a) - \hat{v}_g^* + \max_{a' \in \setname{A}} \hat{q}_b^*(s', a', \vecb{w})
        - \hat{q}_b^*(s, a; \vecb{w})
    }_\text{relative TD in terms of $\hat{q}_b^*$}
    \Big) \nabla \hat{q}_b^*(s, a; \vecb{w})
\Big\},
\label{equ:qbw_update}
\end{equation}
which can be interpreted as the parameterized form of \eqreff{equ:qbstar_update},
and is a \emph{semi-gradient} update, similar to its discounted-reward counterpart
in $Q_\gamma$-learning \citep[\equ{16.3}]{sutton_2018_irl}.
This update is adopted by \citet[\alg{2}]{yang_2016_csv}.
The approximation for $v_g^*$ can be performed in various ways,
as for tabular settings in \secref{sec:valiter_tabular}.
Note that \eqreff{equ:qbw_update} differs from \eqreff{equ:qbw_update_bertsekas}
in the use of TD, affecting the update factor.
In contrast, \citet[\equ{10}]{prashanth_2011_tlrl} leveraged \eqreff{equ:qbstar_update}
in order to suggest the following update,
\begin{equation*}
\vecb{w} \gets \vecb{w} + \beta \Big\{
    r(s, a) - \hat{v}_g^* + \max_{a' \in \setname{A}} \hat{q}_b^*(s', a', \vecb{w}) \Big\},
\quad \text{whose $\hat{v}_g^*$ is updated using \eqreffand{equ:vgstar_sa}{equ:optgain_sa_bertsekas}}.
\end{equation*}

\section{Policy-iteration schemes} \label{sec:politer}

Instead of iterating towards the value of optimal policies, we can iterate
policies directly towards the optimal one.
At each iteration, the current policy iterate is evaluated, then
its value is used to obtain the next policy iterate by taking the greedy action
based on the RHS of the Bellman optimality equation \eqref{equ:avgrew_bellman_optim}.
The latter step differentiates this so-called policy iteration from
naive policy enumeration that evaluates and compares policies as prescribed
in \eqref{equ:gainopt}.

Like in the previous \secref{sec:valiter}, we begin with
the original policy iteration in DP, which is then generalized in~RL.
Afterward, we review average-reward policy gradient methods,
including actor-critic variants, because of their prominence and proven empirical successes.
The last two sections are devoted to approximate policy evaluation, namely
gain and relative value estimations.
\tblref{tbl:politer_work} (in \appref{sec:table_of_existing}) summarizes
existing works on average-reward policy-iteration-based model-free RL.

\subsection{Foundations} \label{sec:politer_backgnd}

In DP, \cite{howard_1960_dpmp} proposed the first policy iteration algorithm to
obtain gain optimal policies for unichain MDPs.
It proceeds as follows.
\begin{enumerate} [label=\textbf{Step~\arabic{*}:}]
\item Initialize the iteration index $k \gets 0$ and
    set the initial policy arbitrarily, $\hat{\pi}^{k=0} \approx \pi^*$.

\item Perform (exact) policy evaluation: \\
    Solve the following underdetermined linear system for $v_g^k$ and $\vecb{v}_b^k$,
    \begin{equation}
        v_b^k(s) + v_g^k = r(s, a) + \sum_{s' \in \setname{S}} p(s' | s, a)\ v_b^k(s'),
        \qquad \forall s \in \setname{S}, \text{with}\ a = \hat{\pi}^k(s),
        \label{equ:poisson}
    \end{equation}
    which is called the Bellman policy expectation equation, also
    the Poisson equation \citep[\equ{9.1}]{feinberg_2002_hmdp}.

\item Perform (exact) policy improvement: \\
    Compute a policy $\hat{\pi}^{k+1}$ by greedy action selection
    (analogous to the RHS of \eqref{equ:avgrew_bellman_optim}):
    \begin{equation*}
        \hat{\pi}^{k+1}(s) \gets \argmax_{a \in \setname{A}} \big[
            \underbrace{
                r(s, a) + \sum_{s' \in \setname{S}} p(s' | s, a)\ v_b^k(s')
            }_\text{$q_b^k(s, a) + v_g^k$}
        \big],\
        \forall s \in \setname{S}.
        \tag{Synchronous updates}
    \end{equation*}

\item If stable, \ie $\hat{\pi}^{k+1}(s) = \hat{\pi}^{k}(s), \forall s \in \setname{S}$,
    then output $\hat{\pi}^{k+1}$ (which is equivalent to $\pi^*$). \\
    Otherwise, increment $k$, then go to Step~2.
\end{enumerate}

The above \emph{exact} policy iteration is generalized to
the generalized policy iteration (GPI) for RL \citep[\secc{4.6}]{sutton_2018_irl}.
Such generalization is in the sense of the details of policy evaluation and improvement,
such as approximation (\emph{inexact} evaluation and \emph{inexact} improvement) and
the update granularity, ranging from per timestep (incremental, step-wise) to
per number (batch) of timesteps.

GPI relies on the policy improvement theorem \citep[\page{101}]{sutton_2018_irl}.
It assures that any $\epsilon$-greedy policy with respect to $q_b(\pi)$ is
an improvement over any $\epsilon$-soft policy $\pi$,
\ie any policy whose
$\pi(a|s) \ge \epsilon / \setsize{A}, \forall (s, a) \in \setname{S} \times \setname{A}$.
Moreover, $\epsilon$-greediness is also beneficial for exploration.
There exist several variations on policy improvement that all share
a similar idea to $\epsilon$-greedy, \ie
updating the current policy \emph{towards} a greedy policy.
They include
\citet[\equ{5}]{wei_2019_avgrew},
\citet[\equ{4}]{abbasi-yadkori_2019_politex},
\citet[\equ{4.1}]{hao_2020_aapi},
and approximate gradient-ascent updates used in policy gradient methods
(\secref{sec:politer_backgnd_polgrad}).

\subsection{Average-reward policy gradient methods} \label{sec:politer_backgnd_polgrad}

In this section, we outline policy gradient methods, which have proven empirical successes
in function approximation settings \citep{agarwal_2019_polgrad, duan_2016_benchrl}.
They requires explicit policy parameterization, which enables
\emph{not only} learning appropriate levels of exploration
(either control- or parameter-based
\citetext{\citealp{vemula_2019_xplor}, \citealp[\fig{1}]{miyamae_2010_npgpe}}),
\emph{but also} injection of domain knowledge \citep[\secc{1.3}]{deisenroth_2013_polsearchrob}.
Note that from a sensitivity-based point of view, policy gradient methods belong to
perturbation analysis \citep[\page{18}]{cao_2007_slo}.

In policy gradient methods, a policy $\pi$ is parameterized by a parameter vector
$\vecb{\theta} \in \Theta = \real{\dim(\vecb{\theta})}$,
where $\dim(\vecb{\theta})$ indicates the number of dimensions of $\vecb{\theta}$.
In order to obtain a smooth dependence on~$\vecb{\theta}$ (hence, smooth gradients),
we restrict the policy class to a set of \emph{randomized} stationary policies~$\piset{SR}$.
We further assume that $\pi(\vecb{\theta})$ is twice differentiable with
bounded first and second derivatives.
For discrete actions, one may utilize a categorical distribution
(a special case of Gibbs/Boltzmann distributions) as follows,
\begin{equation*}
\pi(a | s; \vecb{\theta}) \eqdef
\frac{\exp(\vecb{\theta}^\intercal \vecb{\phi}(s, a))}{
    \sum_{a' \in \setname{A}} \exp(\vecb{\theta}^\intercal \vecb{\phi}(s, a'))},
\quad \forall (s, a) \in \setname{S} \times \setname{A},
\tag{soft-max in action preferences}
\end{equation*}
where $\vecb{\phi}(s, a)$ is the feature vector of a state-action pair $(s, a)$
for this policy parameterization
(\citet[\secc{3.1}]{sutton_2018_irl}, \citet[\equ{8}]{bhatnagar_2009_nac}).
Note that parametric policies may not contain the optimal policy because
there are typically fewer parameters than state-action pairs, yielding some approximation error.

The policy improvement is based on the following optimization with
$v_g(\vecb{\theta}) \eqdef v_g(\pi(\vecb{\theta}))$,
\begin{equation}
\vecb{\theta}^* \eqdef \argmax_{\vecb{\theta} \in \Theta} v_g(\vecb{\theta}),
\quad \text{with iterative update:}\
\vecb{\theta} \gets \vecb{\theta}
+ \alpha \mat{C}^{-1} \nabla v_g(\vecb{\theta}),
\label{equ:polparam_update}
\end{equation}
where $\alpha$ is a positive step length, and
$\mat{C} \in \real{\dim(\vecb{\theta}) \times \dim(\vecb{\theta}) }$ denotes
some preconditioning positive definite matrix.
Based on \eqreff{equ:gain_prob}, we have
\begin{align}
\nabla v_g(\vecb{\theta})
& = \sum_{s \in \setname{S}} \sum_{a \in \setname{A}}
    r(s, a)\ \nabla \{ p_{\pi}^\star(s)\ \pi(a|s; \vecb{\theta}) \}
    \tag{$\nabla \eqdef \partdiff{\vecb{\theta}}$} \\
& = \sum_{s \in \setname{S}} \sum_{a \in \setname{A}}
    p_{\pi}^\star(s)\ \pi(a|s; \vecb{\theta})\ r(s, a)\
    \{ \nabla \log \pi(a|s; \vecb{\theta}) + \nabla \log p_{\pi}^\star(s) \}
    \notag \\ %
& = \sum_{s \in \setname{S}} \sum_{a \in \setname{A}}
        p_{\pi}^\star(s)\ \pi(a|s; \vecb{\theta})\
        \underbrace{q_b^\pi(s, a)}_\text{in lieu of $r(s, a)$}
        \nabla \log \pi(a|s; \vecb{\theta})
    \tag{does not involve $\nabla \log p_{\pi}^\star(s)$} \\
& = \sum_{s \in \setname{S}} \sum_{a \in \setname{A}} \sum_{s' \in \setname{S}}
    p_{\pi}^\star(s)\ \pi(a|s; \vecb{\theta})\ p(s'| s, a)\
    v_b^\pi(s')\ \nabla \log \pi(a|s; \vecb{\theta}).
\label{equ:polgrad}
\end{align}
The penultimate equation above is due to the (randomized) policy gradient theorem
(\citet[\thm{1}]{sutton_2000_pgfnapprox}, \citet[\secc{6}]{marbach_2001_simopt},
\citet[\page{28}]{deisenroth_2013_polsearchrob}).
The last equation was proven to be equivalent by \citet[\app{B}]{castro_2010_stsac}.

There exist (at least) two variants of preconditioning matrices $\mat{C}$.
\textbf{First} is through the second derivative
$\mat{C} = - \nabla^2 v_g(\vecb{\theta})$, as well as its approximation,
see \citet[\app{B.1}]{furmston_2016_newtonps},
\cite[\secc{3.3, 5.2}]{morimura_2008_nnpg}, \citet[\equ{6}]{kakade_2002_npg}.
\textbf{Second}, one can use a Riemannian-metric matrix for natural gradients.
It aims to make the update directions invariant to the policy parameterization.
\cite{kakade_2002_npg} first proposed the Fisher information matrix (FIM) as such a matrix,
for which an incremental approximation was suggested by \citet[\equ{26}]{bhatnagar_2009_nac}.
A generalized variant of natural gradients was introduced by
\cite{morimura_2009_gnac, morimura_2008_nnpg}.
In addition, \cite{thomas_2014_genga} derived another generalization that
allows for a positive semidefinite matrix.
Recall that FIM is only guaranteed to be positive semidefinite
(hence, describing a semi-Riemannian manifold);
whereas the natural gradient ascent assumes the function being optimized
has a Riemannian manifold domain. %

In order to obtain more efficient learning with efficient computation,
several works propose using backward-view eligibility traces in policy parameter updates.
The key idea is to keep track of the eligibility of each component of $\vecb{\theta}$
for getting updated whenever a reinforcing event, \ie $q_b^\pi$, occurs.
Given a state-action sample $(s, a)$, the update in \eqref{equ:polparam_update}
becomes
\begin{equation}
\vecb{\theta} \gets \vecb{\theta}
+ \alpha \mat{C}^{-1} q_b^\pi(s, a)\ \vecb{e}_{\vecb{\theta}},
\quad \text{which is carried out after}\
\vecb{e}_{\vecb{\theta}} \gets
    \lambda_{\vecb{\theta}} \vecb{e}_{\vecb{\theta}}
    + \underbrace{\nabla \log \pi(a|s; \vecb{\theta})}_\text{the eligibility},
\label{equ:theta_update}
\end{equation}
where $\vecb{e}_{\vecb{\theta}} \in \real{\dim(\vecb{\theta})}$ denotes
the accumulating eligibility vector for $\vecb{\theta}$ and
is initialized to $\vecb{0}$, whereas
$\lambda_{\vecb{\theta}} \in (0, 1)$ the trace decay factor for $\vecb{\theta}$.
This is used by
\citet[\secc{4}]{iwaki_2019_i2nac}, %
\citet[\secc{13.6}]{sutton_2018_irl}, %
\citet[\app{B.1}]{furmston_2016_newtonps}, %
\citet[\secc{III.B}]{degris_2012_contact}, %
\citet[\secc{4}]{matsubara_2010_lrpg}, and %
\citet[\secc{5}]{marbach_2001_simopt}. %

As can be observed in \eqreff{equ:polgrad}, computing the gradient
$\nabla v_g(\vecb{\theta})$ exactly
requires (exact) knowledge of $p_{\pi}^\star$ and $q_b^\pi$,
which are both unknown in RL.
It also requires summation over all states and actions, which becomes an issue
when state and action sets are large.
As a result, we resort to the gradient estimate, which if \emph{unbiased},
leads to \emph{stochastic} gradient ascent.

In order to reduce the variance of gradient estimates, there are two common techniques
based on control variates \citep[\secc{5, 6}]{greensmith_2004_varredgrad}.

\textbf{First} is the baseline control variate, for which the \emph{optimal} baseline is
the one that minimizes the variance.
Choosing $v_b^\pi$ as a baseline \citep[\lmm{2}]{bhatnagar_2009_nac}
yields
\begin{equation}
\underbrace{
    q_b^\pi(s, a) - v_b^\pi(s)
}_\text{relative action advantage, $\adv_b^\pi(s, a)$}
= \mathbb{E}_{S'} \Big[
    \underbrace{
        \Big\{ r(s, a) - v_g^\pi + v_b^\pi(S') \Big\}
        - v_b^\pi(s) \big| s, a
    }_\text{relative TD, $\delta_{v_b}^\pi(s, a, S')$}
\Big],
\quad \forall (s, a) \in \setname{S} \times \setname{A},
\label{equ:td_vs_adv}
\end{equation}
where $S'$ denotes the next state given the current state~$s$ and action~$a$.
It has been shown that
$\adv_b^\pi(s, a) = \E{S'}{\delta_{v_b}^\pi(s, a, S')}$,
meaning that the TD, \ie $\delta_{v_b}^\pi(s, a, S')$, can be used as
an \emph{unbiased} estimate for the action advantage
\citetext{\citealp[\prop{1}]{iwaki_2019_i2nac}, \citealp[\thm{5}]{castro_2010_stsac},
\citealp[\lmm{3}]{bhatnagar_2009_nac}}.
Hence, we have
\begin{equation*}
\nabla v_g(\vecb{\theta})
= \mathbb{E}_{S, A} \big[
    \adv_b^\pi(S, A)\ \nabla \log \pi(A|S; \vecb{\theta}) \big]
\approx \delta_{v_b}^\pi(s, a, s')\ \nabla \log \pi(a|s; \vecb{\theta}),
\end{equation*}
where the expectation of $S$, $A$, and $S'$ is approximated with a single sample $(s, a, s')$.
This yields an unbiased gradient estimate with lower variance than that using
$q_b^\pi$.
Note that in RL, the exact $\adv_b^\pi$ and $v_b^\pi$
(for calculating $\delta_{v_b}^\pi$) should also be approximated.

\textbf{Second}, a policy value estimator is set up and often also parameterized by
$\vecb{w} \in \setname{W} = \real{\dim(\vecb{w})}$.
This gives rise to actor-critic methods, where ``actor'' refers to the parameterized policy,
whereas ``critic'' the parameterized value estimator.
The critic can take either one of these three forms,
\begin{enumerate} [label=\roman{*}.]
\item relative state-value estimator $\hat{v}_b^{\pi}(s; \vecb{w}_v)$
    for computing the relative TD approximate $\hat{\delta}_{v_b}^\pi$,
\item both relative state- and action-value estimators for
    $\hat{\adv}_b^\pi(s, a) \gets
    \hat{q}_b^\pi(s, a; \vecb{w}_q) - \hat{v}_b^{\pi}(s; \vecb{w}_v)$,
\item relative action-advantage estimator $\hat{\adv}_b^\pi(s, a; \vecb{w}_{\adv})$,
\end{enumerate}
for all $s \in \setname{S}, a \in \setname{A}$, and
with the corresponding critic parameter vectors $\vecb{w}_v$, $\vecb{w}_q$, and $\vecb{w}_{\adv}$.
This parametric value approximation is reviewed in \secref{sec:politer_valueapprox}.

\subsection{Gain Approximation} \label{sec:politer_gainapprox}

We classify gain approximations into incremental and batch categories.
Incremental methods update their current estimates using information
only from one timestep, hence step-wise updates.
In contrast, batch methods uses information from multiple timesteps in a batch.
Every update, therefore, has to wait until the batch is available.
Notationally, we write $g^\pi \eqdef v_g(\pi)$.
The superscript $\pi$ is often dropped whenever the context is clear.

\subsubsection{Incremental methods}

Based on \eqreff{equ:gain_lim2}, we can define an increasing function
$f(g) \eqdef g - \E{}{r(S, A)}$.
Therefore, the problem of estimating $g$ becomes finding the root of $f(g)$,
by which $f(g) = 0$.
As can be observed, $f(g)$ is unknown because $\E{}{r(S, A)}$ is unknown.
Moreover, we can only obtain noisy observations of $\E{}{r(S, A)}$, namely
$r(s, a) = \E{}{r(S, A)} + \varepsilon$ for some additive error $\varepsilon > 0$.
Thus, the \emph{noisy} observation of $f(\hat{g}_t)$ at iteration $t$ is given by
$\hat{f}(\hat{g}_t) = \hat{g}_t - r(s_t, a_t)$.
Recall that an increasing function satisfies $d f(x)/ dx > 0$ for $x > x^*$
where $x^*$ indicates the root of the $f(x)$.

We can solve for the root of $f(g)$ iteratively via
the Robbins-Monro (RM) algorithm as follows, %
\begin{align}
\hat{g}_{t+1}
& = \hat{g}_t - \beta_t \{ \hat{f}(\hat{g}_t) \}
    = \hat{g}_t - \beta_t \{ \hat{g}_t - r(s_t, a_t) \}
    \tag{for an increasing function $f$} \\
& = \hat{g}_t + \beta_t \{ r(s_t, a_t) - \hat{g}_t \}
= (1 - \beta_t) \hat{g}_t + \beta_t \{ r(s_t, a_t) \},
\label{equ:gain_basic}
\end{align}
where $\beta_t \in (0, 1)$ is a gain approximation step size;
note that setting $\beta_t \ge 1$ violates the RM algorithm since
it makes the coefficient of $\hat{g}_t$ non-positive.
This recursive procedure converges to the root of $f(g)$ with probability~1
under several conditions, including $\varepsilon_t$ is i.i.d with
$\E{}{\varepsilon_t} = 0$,
the step size satisfies the standard requirement in SA, and
some other technical conditions \citep[\secc{6.1}]{cao_2007_slo}.
The stochastic approximation technique in \eqreff{equ:gain_basic} is
the most commonly used.
For instance, \citet[\alg{1}]{wu_2020_ttsac},
\cite{heess_2012_acrl}, \citet[\secc{10.7}]{powell_2011_adp},
\citet[\equ{13}]{castro_2010_stsac}, \citet[\equ{21}]{bhatnagar_2009_nac},
\citet[\equ{3.1}]{konda_2003_aca}, \citet[\secc{4}]{marbach_2001_simopt}, and
\citet[\secc{2}]{tsitsiklis_1999_avgtd}.

Furthermore, a learning rate of $\beta_t = 1/(t + 1)$ in \eqreff{equ:gain_basic}
yields
\begin{equation}
\hat{g}_{t+1}
= \frac{t \times \hat{g}_t + r(s_t, a_t)}{t+1}
= \frac{r(s_t, a_t) + r(s_{t - 1}, a_{t - 1}) + \ldots + r(s_0, a_0)}{t + 1},
\label{equ:lr_special_sa}
\end{equation}
which is the average of a noisy gain observation sequence up to $t$.
The ergodic theorem asserts that for such a Markovian sequence, the time average
converges to \eqref{equ:gain_lim2} as $t$ approaches infinity
\citep[\ch{8}]{gray_2009_prob}.
This decaying learning rate is suggested by \citet[\alg{2}]{singh_1994_avgrewrl}.
Additionally, a variant of $\beta_t = 1/(t + 1)^\kappa$ with
a positive constant $\kappa < 1$
is used by \citet[\secc{5.2.1}]{wu_2020_ttsac} for establishing
the finite-time error rate of this gain approximation under non-i.i.d Markovian samples.

Another iterative approach is based on the Bellman expectation equation \eqref{equ:poisson}
for obtaining the noisy observation of $g$.
That is,
\begin{align}
\hat{g}_{t + 1}
& = (1 - \beta_t) \hat{g}_t
    + \beta_t \big\{
        \underbrace{
            r(s_t, a_t) + \hat{v}_{b, t}^\pi(s_{t+1}) - \hat{v}_{b, t}^\pi(s_t)
        }_\text{$g$ in expectation of $S_{t+1}$ when $\hat{v}_{b, t}^\pi = v_b^\pi$}
    \big\} \tag{The variance is reduced, \cf \eqreff{equ:gain_basic}} \\
& = \hat{g}_t
    + \beta_t \big\{
        \underbrace{
            {\color{red}\{}
                r(s_t, a_t) - \hat{g}_t + \hat{v}_{b, t}^\pi(s_{t+1})
            {\color{red}\}}
            - \hat{v}_{b, t}^\pi(s_t)
        }_\text{$\hat{\delta}_{v_b, t}^\pi(s_t, a_t, s_{t+1})$}
    \big\}.
\label{equ:gain_tdbased}
\end{align}
In comparison to \eqreff{equ:gain_basic}, the preceding update is anticipated to
have lower variance due to the adjustment terms of
$\hat{v}_{b, t}^\pi(s_{t+1}) - \hat{v}_{b, t}^\pi(s_t)$.
This update is used by \citet[\alg{1}]{singh_1994_avgrewrl}, \citet[\secc{3B}]{degris_2012_contact},
and \citet[\page{333}]{sutton_2018_irl}.
An analogous update can be formulated by replacing $v_b^\pi$ with $q_b^\pi$,
which is possible only when the next action $a_{t+1}$ is already available,
as in the differential Sarsa algorithm \citep[\page{251}]{sutton_2018_irl}.

\subsubsection{Batch methods}

In batch settings, there are at least two ways to approximate the gain of a policy,
denoted as $g$.

\textbf{First,} $\hat{g}$ is set to the total rewards \emph{divided} by
the number of timesteps in a batch, as in \eqref{equ:lr_special_sa}, which
also shows its incremental equivalence.
This is used in
\citetext{\citealp[\page{339}]{powell_2011_adp}; \citealp[\secc{2B}]{yu_2009_lspe};
\citealp[\secc{4}]{gosavi_2004_qplearn}, \citealp[\page{319}]{feinberg_2002_hmdp}}.
The work of \citet[\equ{16}]{marbach_2001_simopt} also falls in this category,
but $\hat{g}$ is computed as a weighted average of \emph{all} past rewards from
all batches (regenerative cycles), mixing on- and off-policy samples.
Nonetheless, it is shown to yield lower estimation variance.

\textbf{Second,} based on similiar justification as RVI, we can specify
a reference state and action pair.
For instance,
\citet[\equ{6.23}, \page{311}]{cao_2007_slo} proposed $\hat{g} \gets \hat{v}_b^\pi(\sref)$,
whereas
\citet[\app{A}]{lagoudakis_2003_thesis} advocated $\hat{g} \gets \hat{q}_b^\pi(\sref, \aref)$.

\subsection{Relative state- and action-value approximation}
\label{sec:politer_valueapprox}

\subsubsection{Tabular methods} \label{sec:politer_vapprox_tab}

\citet[\alg{1, 2}]{singh_1994_avgrewrl} proposed the following incremental TD-based update,
\begin{equation*}
\hat{v}_b^\pi(s) \gets \hat{v}_b^\pi(s)
+ \beta \{
    \underbrace{
        \big( r(s, a) - \hat{v}_g^\pi + \hat{v}_b^\pi(s')  \big) - \hat{v}_b^\pi(s)
    }_{\delta_{v_b}^\pi(s, a, s')} \},
\quad \text{for a sample $(s, a, s')$}.
\end{equation*}
A similar update for $\hat{q}_b^\pi$ can be obtained by substituting
$\hat{v}_b^\pi$ with $\hat{q}_b^\pi$ along with a sample for the next action,
as in Sarsa-like algorithms.

In a batch fashion, the relative action value can be approximated as follows,
\begin{equation}
q_b^\pi(s_t, a_t) \approx
\underbrace{
    \sum_{\tau = t}^{t_{\sref}^\pi} r(s_\tau, a_\tau) - \hat{v}_g^\pi,
}_\text{an episode return}
\quad \text{using a sample set $\{(s_\tau, a_\tau) \}_{\tau = t}^{t_{\sref}^\pi}$}
\label{equ:qb_batch_return}
\end{equation}
where $t_{\sref}^\pi$ denotes the timestep at which $\sref$ is visited while
following a policy~$\pi$, assuming that $\sref$ is a recurrent state under all policies.
This is used by \citet[\secc{1}]{sutton_2000_pgfnapprox}, and \citet[\secc{6}]{marbach_2001_simopt}.
Another batch approximation technique is based on the inverse-propensity scoring
\citep[\alg{3}]{wei_2019_avgrew}. %

\subsubsection{Methods with function approximation} \label{sec:politer_vapprox_fa}

Here, we review gradient-based approximate policy evaluation that relies on
the gradient of an error function to update its parameter.
Note that there exist approximation techniques based on least squares, \eg
LSPE($\lambda$) \citep{yu_2009_lspe},
gLSTD and LSTDc \citep{ueno_2008_lstd}, and
LSTD-Q \citep[\app{A}]{lagoudakis_2003_thesis}.

\paragraph{Relative state-value approximation:}
The value approximator is parameterized as
$\hat{\vecb{v}}_b^\pi(\vecb{w}) \approx \vecb{v}_b^\pi$,
where ${\vecb{w} \in \setname{W} = \real{\dim(\vecb{w})}}$.
Then, the mean squared error (MSE) objective is minimized.
That is,
\begin{equation*}
\| \vecb{v}_b^\pi - \hat{\vecb{v}}_b^\pi(\vecb{w}) \|_{\diag(\vecb{p}_\pi^\star)}^2
\eqdef \E{S \sim p_\pi^\star}{\{ v_b^\pi(S) - \hat{v}_b^\pi(S, \vecb{w}) \}^2 },
\end{equation*}
where $\diag(\vecb{p}_\pi^\star)$ is a $\setsize{S}$-by-$\setsize{S}$ diagonal matrix with
$\vecb{p}_\pi^\star(s)$ in its diagonal.
As a result, the gradient estimate for updating $\vecb{w}$ is given by
\begin{align}
- \frac{1}{2} \nabla \E{S \sim p_\pi^\star}{\{ v_b^\pi(S) - \hat{v}_b^\pi(S; \vecb{w}) \}^2 }
& = \E{S \sim p_\pi^\star}{
        \{ v_b^\pi(S) - \hat{v}_b^\pi(S; \vecb{w})  \}
        \nabla \hat{v}_b^\pi(S; \vecb{w})}
    \notag \\
& \approx \{ v_b^\pi(s) - \hat{v}_b^\pi(s; \vecb{w})  \}
        \nabla \hat{v}_b^\pi(s; \vecb{w})
    \tag{Via a single sample $s$} \\
& \approx \big\{
        \underbrace{
            {\color{red} \{}
                r(s, a) - \hat{v}_g^\pi + \hat{v}_b^\pi(s'; \vecb{w})
            {\color{red} \}}
        }_\text{TD target that approximates $v_b^\pi(s)$}
        -\ \hat{v}_b^\pi(s; \vecb{w})
    \big\} \nabla \hat{v}_b^\pi(s; \vecb{w})
    \notag \\
& = \hat{\delta}_{v_b}^\pi(s, a, s'; \vecb{w}) \nabla \hat{v}_b^\pi(s; \vecb{w}).
\label{equ:v_weight_update}
\end{align}
The key difference to supervised learning lies in the fact that
the true value $v_b^\pi(s)$ is approximated by a bootstrapping TD target
involving $\hat{v}_b^\pi(s'; \vecb{w})$, which is biased due to its dependency on $\vecb{w}$.
Such a dependency, however, is not captured (hence, ignored) in the gradient in \eqreff{equ:v_weight_update}
because the TD target plays the role of true values.
As a result, this way of learning $\vecb{w}$ belongs to semi-gradient methods.

With the same motivation as in policy gradient updates \eqreff{equ:theta_update},
one may use an (accumulating) eligibility trace vector $\vecb{e}_v$
with reinforcing event $\delta_{v_b}^\pi$.
This leads to a backward view of TD($\lambda$), whose weight update is given by
\begin{equation}
\vecb{w}_{t+1}
= \vecb{w}_t + \beta_t \hat{\delta}_{v_b}^\pi(s_t, a_t, s_{t+1}) \vecb{e}_{v, t},
\qquad \text{with}\
\vecb{e}_{v, t} = \lambda_v \vecb{e}_{v, t - 1} + \nabla \hat{v}_b^\pi(s_t; \vecb{w}_t),
\label{equ:wv_update}
\end{equation}
for a trace decay factor $\lambda_v \in (0, 1)$ and $\vecb{e}_{-1} = \vecb{0}$
such that
$\vecb{e}_{v, t} =
\sum_{\tau = 0}^t \lambda_v^{t - \tau} \nabla \hat{v}_b^\pi(s_\tau; \vecb{w}_\tau)$.
For linear function approximation, \citet[\thm{1}]{tsitsiklis_1999_avgtd} provide
an asymptotic convergence proof (with probability 1 with incremental gain estimation,
as well as with fixed gain estimates)
and a bound on the resulting approximation error.
This TD($\lambda$) learning for $\hat{v}_b^\pi$ is also used by
\citet[\equ{11.4}, \page{333}]{sutton_2018_irl}, and
\citet[\equ{13}]{castro_2010_stsac}.

As can be observed, there are 2 dedicated step lengths,
\ie the actor's $\alpha_t$ \eqreff{equ:theta_update} and
the critic's $\beta_t$ \eqreff{equ:wv_update}.
Together, they form 2-timescale SA-type approximation,
where both actor and critic are updated at each iteration
but with different step lengths, $\alpha_t \ne \beta_t$.
Intuitively, the policy (actor) should be updated on a slower timescale so that
the critic has enough updates to catch up with the ever-changing policy.
This implies that $\alpha_t$ should be smaller than $\beta_t$.
More precisely, we require that $\lim_{t \to \infty} \beta_t / \alpha_t = \infty$
for assuring the asymptotic convergence \citep{wu_2020_ttsac, bhatnagar_2009_nac, konda_2003_aca}.
Nevertheless, \cite{castro_2010_stsac} argued that from biological standpoint,
2-timescale operation within the same anatomical structure is not well justified.
They analyzed the convergence of 1-timescale actor-critic methods to
the \emph{neighborhood} of a local maximum of $v_g^*$.

\paragraph{Relative action-value and action-advantage approximation:}
Estimation for $q_b^\pi$ can be carried out similarly as that for $v_b^\pi$
in \eqreff{equ:wv_update}, but uses the relative TD on action values, namely
\begin{equation*}
\hat{\delta}_{q_b}^\pi(s, a, s', a') \eqdef
    \underbrace{
        \{ r(s, a) - \hat{v}_g^\pi + \hat{q}_b^\pi(s', a'; \vecb{w}) \}
    }_\text{TD target that approximates $q_b^\pi(s, a)$}
    -\ \hat{q}_b^\pi(s, a; \vecb{w}),
\quad \text{with $a' \eqdef a_{t+1}$},
\end{equation*}
as in Sarsa \citep[\page{251}]{sutton_2018_irl}, as well as \citet[\equ{3.1}]{konda_2003_aca}.
It is also possible to follow this pattern but with different approaches to
approximating the true value of $q_b^\pi(s, a)$, \eg
via episode returns in \eqreff{equ:qb_batch_return} \citep[\thm{2}]{sutton_2000_pgfnapprox}.

For approximating the action advantage parametrically through $\hat{\adv}_b^\pi(\vecb{w})$,
the technique used in \eqreff{equ:v_weight_update} can also be applied.
However, the true value $\mathrm{a}_b^\pi$ is estimated by TD on state values.
That is,
\begin{align*}
- \frac{1}{2} \nabla \E{S \sim p_\pi^\star, A \sim \pi}{
    \{ \adv_b^\pi(S, A) - \hat{\adv}_b^\pi(S, A; \vecb{w}_{\adv}) \}^2 }
& = \E{}{
        \{ \adv_b^\pi(S, A) - \hat{\adv}_b^\pi(S, A; \vecb{w}_{\adv})  \}
        \nabla \hat{\adv}_b^\pi(S, A; \vecb{w}_{\adv})}
    \notag \\
& \approx \big\{ \adv_b^\pi(s, a) - \hat{\adv}_b^\pi(s, a; \vecb{w}_{\adv}) \big\}
        \nabla \hat{\adv}_b^\pi(s, a; \vecb{w}_{\adv})
    \tag{Approximation via a single sample $(s, a)$} \\
& \approx \big\{
    \hat{\delta}_{v_b}^\pi(s, a, s'; \vecb{w}_v)
    -\ \hat{\adv}_b^\pi(s, a; \vecb{w}_{\adv})
    \big\} \nabla \hat{\adv}_b^\pi(s, a; \vecb{w}_{\adv}).
    \tag{Approximation via $\hat{\delta}_{v_b}^\pi \approx \mathrm{a}_b^\pi(s, a)$,
        see \eqreff{equ:td_vs_adv}}
\end{align*}
This is used by \citet[\equ{17}]{iwaki_2019_i2nac}, \citet[\equ{13}]{heess_2012_acrl},
and \citet[\equ{30}]{bhatnagar_2009_nac}.
In particular, \citet[\equ{27}]{iwaki_2019_i2nac} proposed a preconditioning matrix
for the gradients, namely:
$\mat{I} - \kappa (\vecb{f}(s, a) \vecb{f}^\intercal(s, a))/(1 + \kappa \| \vecb{f}(s, a) \|^2)$,
where $\vecb{f}(s, a)$ denotes the feature vector of a state-action pair, while
$\kappa \ge \beta_{\adv}$ some scaling constant.

Futhermore, the action-value or action-advantage estimators can be parameterized linearly
with the so-called $\vecb{\theta}$-compatible state-action feature,
denoted by $\vecb{f}_{\vecb{\theta}}(s, a)$, as follows,
\begin{equation}
\hat{q}_b^\pi(s, a; \vecb{w}) = \vecb{w}^\intercal \vecb{f}_{\vecb{\theta}}(s, a),
\quad \text{with}\
\underbrace{\vecb{f}_{\vecb{\theta}}(s, a) \eqdef \nabla \log \pi(a|s; \vecb{\theta})
    }_\text{$\vecb{\theta}$-compatible state-action feature},
\quad \forall (s, a) \in \setname{S} \times \setname{A}.
\label{equ:compatible_critic}
\end{equation}
This parameterization along with the minimization of MSE loss are beneficial
for two reasons.
\textbf{First}, their use with (locally) optimal parameter $\vecb{w}^* \in \setname{W}$
satisfies
\begin{equation*}
\mathbb{E}_{S, A} \Big[
    \underbrace{\hat{q}_b^\pi(S, A; \vecb{w} = \vecb{w}^*)
        }_\text{in lieu of $q_b^\pi(S, A)$}
    \nabla \pi(A|S; \vecb{\theta})
\Big]
= \nabla v_g(\vecb{\theta}),
\tag{Exact policy gradients when $\vecb{w} = \vecb{w}^*$}
\end{equation*}
as shown by \citet[\thm{2}]{sutton_2000_pgfnapprox}.
Otherwise, the gradient estimate is likely to be biased \citep[\lmm{4}]{bhatnagar_2009_nac}.
\textbf{Second}, they make computing natural gradients equivalent to finding
the optimal weight $\vecb{w}^*$ for $\hat{q}_b^\pi(\vecb{w})$.
That is,
\begin{align}
\sum_{s \in \setname{S}} \sum_{a \in \setname{A}}
p_{\vecb{\theta}}^\star(s) \pi(a|s; \vecb{\theta})
\vecb{f}_{\vecb{\theta}}(s, a) \big(\vecb{f}_{\vecb{\theta}}^\intercal(s, a) \vecb{w}^*
    -  q_b^\pi(s, a) \big)
& = \vecb{0}
    \tag{Since MSE is 0 when $\vecb{w} = \vecb{w}^*$} \\
\Big\{
    \underbrace{
        \sum_{s \in \setname{S}} \sum_{a \in \setname{A}}
        p_{\vecb{\theta}}^\star(s)  \pi(a|s; \vecb{\theta}) \vecb{f}_{\vecb{\theta}}(s, a)
        \vecb{f}_{\vecb{\theta}}^\intercal(s, a)
        }_{\mat{F}_{a}(\vecb{\theta})}
    \Big\} \vecb{w}^*
& = \underbrace{
    \sum_{s \in \setname{S}} \sum_{a \in \setname{A}}
        p_{\vecb{\theta}}^\star(s) \pi(a|s; \vecb{\theta}) q_b^\pi(s, a)
        \vecb{f}_{\vecb{\theta}}(s, a)
    }_{\nabla v_g(\vecb{\theta})} \notag \\
\vecb{w}^* & = \underbrace{
        \mat{F}_{a}^{-1}(\vecb{\theta})\ \nabla v_g(\vecb{\theta})
    }_\text{The natural gradient},
\label{equ:fisher_a}
\end{align}
where $\mat{F}_{a}(\vecb{\theta})$ denotes the Fisher (information) matrix
based on the action distribution (hence, the subscript $a$),
as introduced by \citet[\thm{1}]{kakade_2002_npg}.
This implies that the estimation for natural gradients is reduced to
a regression problem of state-action value functions.
It can be either regressing
\begin{enumerate} [label=\roman{*}.]
\item action value $q_b^\pi$
    \citetext{\citealp[\thm{2}]{sutton_2000_pgfnapprox}, \citealp[\thm{1}]{kakade_2002_npg},
        \citealp[\equ{3.1}]{konda_2003_aca}},
    \label{item:q_regression}
\item action advantage $\adv_b^\pi$
    \citetext{\citealp[\equ{30}, \alg{3, 4}]{bhatnagar_2009_nac};
        \citealp[\equ{13}]{heess_2012_acrl}; \citealp[\equ{17}]{iwaki_2019_i2nac}}, or
    \label{item:a_regression}
\item the immediate reward $r(s, A \sim \pi)$ as a known ground truth
    \citep[\thm{1}]{morimura_2008_nnpg},
    recall that the ground truth $q_b^\pi$ (Item~\ref{item:q_regression}) and
    $\adv_b^\pi$ (Item~\ref{item:a_regression}) above are unknown.
    This reward regression is used along with
    a $\vecb{\theta}$-compatible state-action feature
    that is defined differently compared to \eqref{equ:compatible_critic},
    namely
    $\vecb{f}_{\vecb{\theta}}(s, a) \eqdef \nabla \log p_\pi^\star(s, a)
    = \nabla \log p_\pi^\star(s) + \nabla \log \pi(a|s; \vecb{\theta})$,
    where $\vecb{f}_{\vecb{\theta}}$ is based on
    the stationary joint state-action distribution $p_\pi^\star(s, a)$.
    In fact, this leads to a new Fisher matrix, yielding the so-called
    natural state-action gradients $\mat{F}_{s, a}$
    (which subsumes $\mat{F}_{a}$ in \eqref{equ:fisher_a}
    as a special case when $\nabla \log p_\pi^\star$ is set to $\vecb{0}$).
    \label{item:mori_regression}
\end{enumerate}

In addition, \cite{konda_2003_aca} pointed out that
the dependency of $\nabla v_g(\vecb{\theta})$ to $q_b^\pi$ is only
through its inner products with vectors in the subspace spanned by
$\{ \nabla_i \log \pi(a|s; \vecb{\theta})\}_{i=1}^{\dim(\vecb{\theta})}$.
This implies that learning the projection of $q_b^\pi$ onto the aforementioned
(low-dimensional) subspace is sufficient, instead of learning $q_b^\pi$ fully.
A $\vecb{\theta}$-dependent state-action feature can be defined based on
either the action distribution (as in \eqreff{equ:compatible_critic}), or
the stationary state-action distribution (as mentioned in Item~\ref{item:mori_regression}
in the previous passage).

\section{Discussion} \label{sec:discuss}

In this section, we discuss several open questions as first steps towards
completing the literature in average-reward model-free RL.
We begin with those of value iteration schemes (\secref{sec:discuss_optimalvalue}),
then of policy iteration schemes focussing on policy evaluation
(\secref{sec:discuss_value}).
Lastly, we also outline several issues that apply to both or beyond
the two aforementioned schemes.

\subsection{Approximation for optimal gain and optimal action-values (of optimal policies)}
\label{sec:discuss_optimalvalue}

Two main components of $Q_b$-learning are $\hat{q}_b^*$ and $\hat{v}_g^*$ updates.
The former is typically carried out via either \eqreff{equ:qbstar_update} for tabular,
or \eqreff{equ:qbw_update} for (parametric) function approximation settings.
What remains is determining how to estimate the optimal gain $v_g^*$,
which becomes the main bottleneck for $Q_b$-learning.

As can be seen in \figref{fig:taxonomy_optgainapprox} (\appref{sec:taxonomy}),
there are two classes of $v_g^*$ approximators.
First, approximators that are not SA-based generally need
$\sref$ and $\aref$ specification, which is shown to affect the performance
especially in large state and action sets \citep{wan_2020_avgrew, yang_2016_csv}.
On the other hand, SA-based approximators require a dedicated stepsize $\beta_g$,
yielding more complicated 2-timescale $Q_b$-learning.
Furthermore, it is not yet clear whether the approximation for $v_g^*$ should be on-policy,
\ie updating only when a greedy action is executed, at the cost of reduced sample efficiency.
This begs the question of which approach to estimating $v_g^*$ is ``best'' (in which cases).

In discounted reward settings, \cite{hasselt_2010_doubleq} (also, \cite{hasselt_2016_ddqn})
pointed out that the approximation for $\E{}{\max_{a \in \setname{A}} q_\gamma^*(S_{t+1}, a)}$
poses overestimation, which may be non-uniform and not concentrated at states
that are beneficial in terms of exploration.
He proposed instantiating two decoupled approximators such that
\begin{equation*}
\E{S_{t+1}}{\max_{a' \in \setname{A}} q_\gamma^*(S_{t+1}, a')} \approx
\hat{q}_\gamma^*(s_{t+1}, \argmax_{a' \in \setname{A}} \hat{q}_\gamma^*(s_{t+1}, a';
\vecb{w}_q^{(2)}); \vecb{w}_q^{(1)}),
\tag{Double $Q_\gamma$-learning}
\end{equation*}
where $\vecb{w}_q^{(1)}$ and $\vecb{w}_q^{(2)}$ denote their corresponding weights.
This was shown to be successful in reducing the negative effect of overestimation.
In average reward cases, the overestimation of $q_b^*$ becomes more convoluted
due to the involvement of $\hat{v}_g^*$, as shown in \eqreff{equ:avgrew_bellmanopt_q}.
We believe that it is important to extend the idea of double action-value approximators to $Q_b$-learning.

To our knowledge, there is no finite-time convergence analysis for $Q_b$-learning thus far.
There are also very few works on $Q_b$-learning with function approximation.
This is in contrast with its discounted reward counterpart,
\eg sample complexity of $Q_\gamma$-learning with UCB-exploration bonus \citep{wang_2020_qlearning},
as well as deep $Q_\gamma$-learning neural-network (DQN) and its variants
with \emph{non-linear} function approximation \citep{hessel_2018_rainbow}.

\subsection{Approximation for gain, state-, and action-values of any policy}
\label{sec:discuss_value}

\paragraph{Gain approximation $\hat{v}_g^\pi$:} %
In order to have more flexibility in terms of learning methods, we can parameterize
the gain estimator, for instance,
$\hat{v}_g^\pi(s; \vecb{w}_g) \coloneqq \vecb{f}^\intercal(s) \vecb{w}_g$,
by which the gain \emph{estimate} is state-dependent.
The learning uses the following gradients,
\begin{align*}
-\frac{1}{2} \nabla \E{S \sim p_\pi^\star}{\big\{ v_g^\pi - \hat{v}_g^\pi(S; \vecb{w}_g) \big\}^2}
& = \E{S \sim p_\pi^\star}{\big\{ v_g^\pi - \hat{v}_g^\pi(S; \vecb{w}_g) \big\}
    \nabla \hat{v}_g^\pi(S; \vecb{w}_g)} \\
& \approx \big\{ v_g^\pi(s) - \hat{v}_g^\pi(s; \vecb{w}_g) \big\}
    \nabla \hat{v}_g^\pi(s; \vecb{w}_g)
    \tag{By a single sample $s$} \\
& \approx \big\{
        \underbrace{
            r(s, a) + \hat{v}_b^\pi(s') - \hat{v}_b^\pi(s)
        }_\text{approximates the true $v_g^\pi$ based on \eqreff{equ:poisson}}
        -\ \hat{v}_g^\pi(s; \vecb{w}_g)
    \big\} \nabla \hat{v}_g^\pi(s; \vecb{w}_g) \\
& = \hat{\delta}_{v_b}^\pi(s, a, s'; \vecb{w}_g) \nabla \hat{v}_g^\pi(s; \vecb{w}_g).
\end{align*}
This requires further investigation whether the above parameterized approximator
for the gain $v_g^\pi$ is beneficial in practice.

\paragraph{State-value approximation $\hat{v}_b^\pi$:} %
We observe that most, if not all, relative state-value approximators are based on TD, leading to
semi-gradient methods (\secref{sec:politer_vapprox_fa}).
In discounted reward settings, there are several non-TD approaches.
These include the true gradient methods \citep{sutton_2009_tdc},
as well as those using the Bellman residual (\ie the expected value of TD),
such as \cite{zhang_2019_drrl, geist_2017_bellmanres};
see also surveys by \cite{dann_2014_petd, geist_2013_pvfa}.

It is interesting to investigate whether TD or not TD for learning $\hat{v}_b^\pi$.
For example, we may formulate a (true) gradient TD method that optimizes
the mean-squared \emph{projected} Bellman error (MSPBE), namely
$\| \hat{v}_b^\pi(\vecb{w})
    - \mathbb{P}[\mathbb{B}_g^\pi[\hat{v}_b^\pi(\vecb{w})]] \|_{\diag(\vecb{p}_\pi^\star)}^2$
with a projection operator~$\mathbb{P}$ that projects any value approximation
onto the space of representable paramaterized approximators.

\paragraph{Action-value approximation $\hat{q}_b^\pi$:} %
Commonly, the estimation for $q_b^\pi$ involves the gain estimate $\hat{v}_g^\pi$ as
dictated by the Bellman expectation equation \eqref{equ:poisson}.
It is natural then to ask: \emph{is it possible to perform average-reward GPI
without estimating the gain of a policy at each iteration?}
We speculate that the answer is affirmative as follows.

Let the gain $v_g^\pi$ be the baseline for $q_b^\pi$
in a similiar manner to $v_b^\pi$ in \eqreff{equ:td_vs_adv};
\cf in discounted reward settings, see \citet[\thm{1}]{weaver_2001_orb}.
Then, based on the Bellman expectation equation in action values
(analogous to \eqref{equ:avgrew_bellmanopt_q}), we have the following identity,
\begin{equation*}
q_b^\pi(s, a) - (- v_g^\pi)
= \underbrace{
    r(s, a) + \E{S_{t+1}}{v_b^\pi(S_{t+1})}
}_\text{The surrogate action-value $\mathfrak{q}_b^\pi(s, a)$}.
\tag{Note different symbols, $q_b^\pi$ vs $\mathfrak{q}_b^\pi$}
\end{equation*}
This surrogate can be parameterized as $\hat{\mathfrak{q}}_b^\pi(s, a; \vecb{w})$.
Its parameter $\vecb{w}$ is updated using the following gradient estimates,
\begin{align*}
- \frac{1}{2} \nabla \E{S, A}{\{
    \mathfrak{q}_b^\pi(S, A) - \hat{\mathfrak{q}}_b^\pi(S, A; \vecb{w}) \}^2}
& = \E{S, A}{
    \{ \mathfrak{q}_b^\pi(S, A) - \hat{\mathfrak{q}}_b^\pi(S, A; \vecb{w}) \}
    \nabla \hat{\mathfrak{q}}_b^\pi(S, A; \vecb{w})
    } \\
& = \mathbb{E}_{S, A, S'} \Big[
    \Big\{ \underbrace{r(S, A) + v_b^\pi(S')}_\text{The surrogate $\mathfrak{q}_b^\pi(S, A)$}
    -\ \hat{\mathfrak{q}}_b^\pi(S, A; \vecb{w}) \Big\}
    \nabla \hat{\mathfrak{q}}_b^\pi(S, A; \vecb{w})
    \Big] \\
& \approx \Big\{ r(s, a) + v_b^\pi(s') - \hat{\mathfrak{q}}_b^\pi(s, a; \vecb{w}) \Big\}
    \nabla \hat{\mathfrak{q}}_b^\pi(s, a; \vecb{w})
    \tag{Using a single sample $(s, a, s')$} \\
& \propto \Big\{ r(s, a) +
    \underbrace{v_b^\pi(s') + (v_g^\pi + \kappa)}_\text{The surrogate state-value $\nu_{b}^\pi(s')$}
    -\ \hat{\mathfrak{q}}_b^\pi(s, a; \vecb{w}) \Big\}
    \nabla \hat{\mathfrak{q}}_b^\pi(s, a; \vecb{w}),
    \tag{Note different symbols, $v_b^\pi$ vs $\nu_b^\pi$}
\end{align*}
for some arbitrary constant $\kappa \in \real{}$.
The key is to exploit the fact that we have one degree of freedom in
the underdetermined linear systems for policy evaluation in \eqreff{equ:poisson}.
Here, the surrogate state-value $\nu_{b}^\pi$ is equal to $v_{b}^\pi$ up to
some constant, \ie $\nu_{b}^\pi = v_{b}^\pi + (v_g^\pi + \kappa)$.
One can estimate $\nu_{b}^\pi$ in a similar fashion as $\hat{v}_b^\pi$ in
\eqreff{equ:v_weight_update},
except that now the gain approximation $\hat{v}_g^\pi$ is no longer needed.
The parameterized estimator $\hat{\nu}_{b}^\pi(\vecb{w}_{\nu})$ can be
updated by following below gradient estimate,
\begin{equation}
- \frac{1}{2}\nabla \E{S \sim p_\pi^\star}{\{ \nu_b^\pi(S) - \hat{\nu}_b^\pi(S, \vecb{w}_{\nu}) \}^2 }
\approx \big\{
    r(s, a) + \nu_b^\pi(s'; \vecb{w}_{\nu})
        - \hat{\nu}_b^\pi(s; \vecb{w}_{\nu})
\big\} \nabla \hat{\nu}_b^\pi(s; \vecb{w}_{\nu}),
\label{equ:grad_nu}
\end{equation}
which is based on a single sample $(s, a, s')$.
In the RHS of \eqref{equ:grad_nu} above, the TD of $\nu_b^\pi$ looks similar to
that of discounted reward $v_\gamma$, but with abused $\gamma = 1$.
Therefore, its stability (whether it will grow unboundedly) warrants experiments.

\paragraph{Extras:} %
We notice that existing TD-based value approximators simply re-use the TD estimate
$\hat{\delta}_{v_b}^\pi$ obtained using the \emph{old} gain estimate from the previous iteration,
\eg \citet[\page{251, 333}]{sutton_2018_irl}.
They do not harness the newly updated gain estimate for cheap recomputation of $\hat{\delta}_{v_b}^\pi$.
We anticipate that doing so will yield more accurate evaluation of the current policy.

There are also open research questions that apply to both average- and discounted-reward policy evaluation,
but generally require different treatment.
They are presented below along with the parallel works in discounted rewards, if any,
which are put in brackets at the end of each bullet point.
\begin{itemize}
\item How sensitive is the estimation accuracy of $\hat{v}_b^\pi$ with respect to
    the trace decay factor $\lambda$ in TD($\lambda$)?
    Recall that $\hat{v}_b^\pi$ involves $\hat{v}_g$,
    which is commonly estimated using $\hat{\delta}_{v_b}^\pi$ as in \eqreff{equ:gain_tdbased}.
    (\cf \citet[\fig{12.6}]{sutton_2018_irl})

\item What are the advantages and challenges of natural gradients for the critic?
    Note that natural gradients typically are applied only for the actor.
    (\cf \cite{wu_2017_acktr})

\item As can be observed, the weight update in \eqreff{equ:wv_update} resembles that of SGD with momentum.
    It begs the question: what is the connection between backward-view eligibility traces and momentum
    in gradient-based step-wise updates for both actor and critic parameters?
    (\cf \cite{vieillard_2020_mmtm, nichols_2017_etm, xu_2006_etac})

\item How to construct ``high-performing'' basis vectors for the value approximation?
    To what extent does the limitation of $\vecb{\theta}$-compatible critic
    (\eqreff{equ:compatible_critic}) outweigh its benefit?
    Also notice the Bellman average-reward bases \citep[\secc{11.2.4}]{mahadevan_2009_lrmdp},
    as well as non-linear (over-)parameterized neural networks.
    (\cf \cite{wang_2019_neural})

\item Among 3 choices for advantage approximation (\secref{sec:politer_backgnd_polgrad}),
    which one is most beneficial?
\end{itemize}

\subsection{Further open research questions}

In the following passages, relevant works in discounted reward settings are mentioned,
if any, inside the brackets at the end of each part.

\paragraph{On batch settings:}
For on-policy online RL without experience replay buffer,
we pose the following questions.
How to determine a batch size that balances the trade-off between
collecting more samples with the current policy and
updating the policy more often with fewer numbers of samples?
How to apply the concept of backward-view eligibility traces in batch settings?
(\cf \cite{harb_2017_etdqn}).

\paragraph{On value- vs policy-iteration schemes:}
As can be observed in \tblrefand{tbl:valiter_work}{tbl:politer_work} (in \appref{sec:table_of_existing}),
there is less work on value- than policy-iteration schemes;
even less when considering only function approximation settings.
Our preliminary experiments suggest that although following value iteration is straightforward
(\cf policy iteration with evaluation and improvement steps),
it is more difficult to make it ``work'', especially with function approximation.
One challenge is to specify the proper offset, \eg $v_g^*$ or its estimate,
in RVI-like methods to bound the iterates.
Moreover, \citet[\secc{3.5}]{mahadevan_1996_avgrew} highlighted that
the seminal value-iteration based RL, \ie average reward $Q_b$-learning,
is sensitive to exploration.

We posit these questions.
Is it still worth it to adopt the value iteration scheme after all?
Which scheme is more advantageous in terms of exploration in RL?
How to reconcile both schemes?
(\cf \cite{odonoghue_2016_pgq, schulman_2017_pgql, wang_2020_qlearning}).

\paragraph{On distributional (cf. expected-value) perspective:}
There exist few works on distributional views.
\cite{morimura_2010_sdpg} proposed estimating $\nabla \log p^{\vecb{\theta}}(s)$ for obtaining
the gradient estimate $\hat{\nabla} v_g(\vecb{\theta})$ in \eqreff{equ:polgrad},
removing the need to estimate the action value $q_b^{\vecb{\theta}}$.
One important question is how to scale up the distribution (density) estimation for large RL problems.
(\cf \cite{bellemare_2017_distrib}).

\paragraph{On MDP modeling:}
The broadest class that can be handled by existing average-reward RL is the unichain MDP;
note that most works assume the more specific class, \ie the recurrent (ergodic) MDP.
To our knowledge, there is still no average reward model-free RL for multichain MDPs
(which is the most general class).

We also desire to apply average-reward RL to continuous state problems,
for which we may benefit from the DP theory on general states, \eg \citet[\ch{8}]{sennott_1998_sdp}.
There are few attempts thus far, for instance,
\citep{yang_2019_elqr}, which is limited to linear quadratic regulator with ergodic cost.

\paragraph{On optimality criteria:}
The average reward optimality is underselective for problems with transient states,
for which we need $(n=0)$-discount (bias) optimality, or even higher $n$-discount optimality.
This underselective-ness motivates the weighted optimality in DP \citep{krass_1992_wmdp}.
In RL, \cite{mahadevan_1996_sensitivedo} developed bias-optimal Q-learning.

In the other extreme, $(n=\infty)$-discount optimality is the most selective criterion.
According to \citet[\thm{10.1.5}]{puterman_1994_mdp}, it is proven to be equivalent to
the Blackwell optimality, which intuitively claims that upon considering sufficiently
far into the future via the Blackwell's discount factor $\gammabw$,
there is no policy better than the Blackwell optimal policy.
Moreover, optimizing the discounted reward does not require any knowledge about
the MDP structure (\ie recurrent, unichain, multichain classification).
Therefore, one of pressing questions is on estimating such $\gammabw$ in RL.

\section*{Acknowledgments}
We thank Aaron Snoswell, Nathaniel Du Preez-Wilkinson, Jordan Bishop,
Russell Tsuchida, and Matthew Aitchison
for insightful discussions that helped improve this paper.
Vektor is supported by the University of Queensland Research Training Scholarship.

\bibliography{main}
\bibliographystyle{apalike}

\afterpage{%
\clearpage%
\appendix
\section*{\centering APPENDIX}

\section{Tables of existing works} \label{sec:table_of_existing}
\tblrefand{tbl:valiter_work}{tbl:politer_work} summarize
the major contributions and experiments of the existing works on
average-reward model-free RL that are based on value- and policy-iteration schemes,
respectively.

\section{Taxonomy for approximation techniques} \label{sec:taxonomy}
We present taxonomies of approximation techniques.
They are for optimal gain $v_g^*$ in value-iteration schemes
(\figref{fig:taxonomy_optgainapprox}),
as well as for those in policy-iteration schemes, namely
gain $v_g^\pi$ (\figref{fig:taxonomy_gain}),
state values $v_b^\pi$ (\figref{fig:taxonomy_stateval}),
and action-related values $q_b^\pi$ and $\adv_b^\pi$ (\figref{fig:taxonomy_actval}).

\section{Existing benchmarking environments} \label{sec:problem_spec}

Different average-reward RL methods are often evaluated with different sets of environments,
making comparisons difficult.
Therefore, it is imperative to recapitulate those environments in order to
calibrate the progress, as well as
for a sanity check (before targeting more complex environments).

We compile a variety of environments that are used in existing works.
They are classified into two categories, namely continuing (non-episodic)
and episodic environments.

\subsection{Continuing environments}
Continuing environments are those with no terminating goal state and
hence continue on infinitely.
These environments induce infinite horizon MDPs and are typically geared towards
everlasting real-world domains found in areas such as business,
stock market decision making, manufacturing, power management,
traffic control, server queueing operation and communication networks.

There are various different continuous environments used in average-reward RL literature
with the more popular detailed below.
Other examples include
Preventive Maintenance in Product Inventory \citep{das_1999_smart},
Optimal Portfolio Choice \citep{ormoneit_2001_kbrl},
Obstacle avoidance task \citep{mahadevan_1996_avgrew}, and
Multi-Armed Restless Bandits \citep{avrachenkov_2020_wiql}.

\subsubsection{Symbolic MDPs}

In order to initially test the performance of algorithms in RL, simple environments are
required to use as a test bed. These environments often serve no practical purpose and
are developed purely for testing. Below we describe three such environments being $n$-state,
$n$-cycle and $n$-chain.

\paragraph{$n$-state (randomly-constructed) MDPs:} \mbox{} \\
PI: \cite{singh_1994_avgrewrl, kakade_2002_npg, morimura_2008_nnpg, morimura_2009_gnac, bhatnagar_2009_nactr, castro_2010_stsac, matsubara_2010_asspg, morimura_2014_pgrl, wei_2019_avgrew, hao_2020_aapi}. \\
VI: \cite{schwartz_1993_rlearn, singh_1994_avgrewrl, jafarniajahromi_2020_nor}.\\

These are environments in which an MDPs with $n$ states is constructed with
randomized transition matrices and rewards. In order to provide context
and meaning, a background story is often provided for these systems such as in the case of
DeepSea \citep{hao_2020_aapi} where the environment is interpreted
as a diver searching for treasure.

One special subclass is the Generic Average Reward Non-stationary Environment Testbed (GARNET),
which was proposed by \cite{bhatnagar_2009_nactr}.
It is parameterized as GARNET($\setsize{S}, \setsize{A}, x,\sigma,\tau$),
where $\setsize{S}$ is the number of states and $\setsize{A}$ is the number of actions,
$\tau$ determines how non-stationary the problem is, and
$x$ is a branching factors which determines how many next states are available
for each state-action pair.
The set of next states is created by a random selection from the state set without replacement.
Probabilities of going to each available next state are uniformly distributed and the reward
recieved for each transition is normally distributed with standard deviation $\sigma$.

\paragraph{$n$-cycle (loop, circle) MDPs:} \mbox{} \\
PI: \emph{None}. \\
VI: \cite{schwartz_1993_rlearn, mahadevan_1996_avgrew, yang_2016_csv, wan_2020_avgrew}.\\

This group of problem has $n$ cycles (circles, loops) with the state set
defined by the number of cycles and the number of states in each cycle.
The state transitions are typically deterministic.
At the state in which the loops intersect, the agent must decide which loop to take.
For each state inside the loops, the only available actions are to stay or move to the
next state.
The reward function is set out such that each loop leads to a different reward,
with some rewards being more delayed than others.

\cite{mahadevan_1996_avgrew} contains a 2-cycle problem. A robot receives a reward of
+5 if it chooses the cycle going from home to the printer which has 5 states. It receives
+20 if it chooses the other cycle which leads to the mail room and contains 10 states.

\paragraph{$n$-chain:} \mbox{} \\
PI: \cite{wei_2019_avgrew}. \\
VI: \cite{yang_2016_csv, jafarniajahromi_2020_nor}.

These are environments involving $n$ states arranged in a linear chain-like formation.
The only two available actions are to move forwards or backwards.
Note that the term ``chain'' here does not mean a recurrent class.
All $n$-chain environments has one recurrent class.

\cite{strens_2000_bfrl} presents are more specific type which is used in the OpenAI gym.
In this, moving forward results in no reward whereas choosing to go backwards returns
the agent to beginning results in a small reward.
Reaching the end of the chain presents a large reward.
The state transitions are not deterministic, with each action having a small possibility
of resulting in the opposite transition.

RiverSwim is another $n$-chain style problem consisting of 6 states in a chain-like formation to
represent positions across a river with a current flowing from right to left.
Each state has two possible actions; swim left with the current or right against
the current. Swimming with the current is always successful (deterministic) but swimming against
it can fail with some probability. The reward function is such that the
optimal policy is swimming to the rightmost state and remaining there.

\subsubsection{Queuing}

Another commonly used group of environments in RL involve optimization of systems that
involve queuing.
Common examples of queuing environments within literature include call bandwidth
allocation \citep{marbach_2001_simopt}, and seating on airplanes
\citep{gosavi_2002_smart}.

\paragraph{Server-Access Control Queue:} \mbox{} \\
PI: \cite{sutton_2018_irl}: \page{252}, \cite{devra_2018_dtdl, abbasi-yadkori_2019_politex, abbasi-yadkori_2019_xpert} \\
VI: \cite{wan_2020_avgrew, schneckenreither_2020_avgrew}.

The problem of controlling the access of queuing customers to a limited number of servers.
Each customer belongs to a priority group and the state of the system (MDP) is
described by this priority as well as the number of free servers.
At each time-step, the servers become available with some probability and the agent can
accept or reject service to the first customer in each line.
Accepting a customer gives a reward proportional to the priority
of that customer when their service is complete.
Rejecting a customer always results in a reward of 0.
The goal of this problem is to maximize the reward received by making decisions based on
customer priority and the number of available servers.

\paragraph{Traffic Signal Control:} \mbox{} \\
PI: \cite{prashanth_2011_tlrl, liu_2018_offpol} \\
VI: \cite{prashanth_2011_tlrl}

Maximizing the flow of traffic across a network of junctions through finding the optimal
traffic signal configuration.
The controller periodically receives information about congestion which gives us a state
consisting of a vector containing information of queue length and elapsed time since each lane
turned red.
Actions available are the sign configurations of signals which can be turned to green simultaneously.
The cost (negative reward) is the sum of all queue lengths and elapsed times across the whole network.
Queue length is used to reduce congestion and the elapsed time ensures fairness for all lanes.

\subsubsection{Continuous States and Actions}

Many environments with applications to real world systems such as robotics can contain
continuous states and actions.
The most common approach to these environments in RL is to discretize the continuous variables
however there are RL methods that can be used even if the MDP remains continuous.
Below we describe commonly used continuous environments from within the literature.

\paragraph{Linear Quadratic Regulators:} \mbox{} \\
PI: \cite{kakade_2002_npg}. \\
VI: \emph{None}.

The Linear Quadratic Regulator (LQR) problem is fundamental in RL due to the
simple structure providing a useful tool for assessing different methods.
The systems dynamics are $s_{t+1} = \mat{A} s_t + \mat{B} a_t + \epsilon_t$
where $\epsilon_t$ is uniformly distributed random noise, that is i.i.d.
for each $t \geq 0$
The cost (negative reward) function for the system is $c(s,a)=s^\intercal Qx + a^\intercal Ra$.
$\mat{A}$, $\mat{B}$, $\mat{Q}$ and $\mat{R}$ are all matrices with proper dimensions.
This problem is viewed as an MDP with
state and action spaces of $\setname{S} = \real{d}$ and $\setname{A} = \real{k}$ respectively.
The optimal policy of LQR is a linear function of the state.

\paragraph{Swing-up pendulum:} \mbox{} \\
PI: \cite{morimura_2010_sdpg, degris_2012_offpolac, liu_2018_offpol}. \\
VI: \emph{None}.

Swinging a simulated pendulum with the goal of getting it into
and keeping it in the vertical position.
The state of the system is described by the current angle and angular
velocity of the pendulum.
The action is the torque applied at the base of the pendulum
which is restricted to be within a practical limit.
Rewards are received as the cosine of pendulums angle which is
proportional to the height of the pendulum's tip.

\subsection{Episodic environments}

An episodic problem has at least one terminal state
(note that multiple types of terminals can be modeled as one terminal state when
the model concerns only with $r(s, a)$, instead of $r(s, a, s')$).
Once an agent enters the terminal state, the agent-environment interaction terminates.
An episode refers to a sequence of states, actions, and rewards from $t = 0$
until entering the terminal state.
It is typically assumed that the termination eventually occurs in finite time.

\paragraph{Grid-navigation:} \mbox{} \\
PI: \emph{None}. \\
VI: \cite{mahadevan_1996_avgrew}.

\paragraph{Tetris:} \mbox{} \\
PI: \cite{kakade_2002_npg}. \\
VI: \cite{yang_2016_csv}.

\paragraph{Atari Pacman:} \mbox{} \\
PI: \cite{abbasi-yadkori_2019_politex}. \\
VI: \emph{None}.

\afterpage{%
\clearpage%
\begin{figure}
\begin{tikzpicture}[
rootnode/.style = {
    shape = rectangle, rounded corners,
    fill           = blue!40!white,
    minimum width  = 1cm,
    minimum height = 1cm,
    align          = center,
    text           = white},
nonleafnode/.style = {
    shape = rectangle, rounded corners,
    fill           = blue!40!white,
    minimum width  = 1.5cm,
    minimum height = 1cm,
    align          = center,
    text           = white},
leafnode/.style = {
    shape = rectangle,
    fill           = white,
    minimum width  = 3cm,
    minimum height = 1.5cm,
    align          = left,
    text           = black},
edge/.style  = {
    -,
    ultra thick,
    blue!40!white,
    shorten >= 4pt}
]

\node[rootnode](0;0) at (0, 0) {$\hat{v}_g^*$};
\node[nonleafnode](1;0) at (3, 4) {SA \\ (2-timescale $Q_b$-learning)};
    \node[nonleafnode](2;00) at (7.5,8) {
        Update only when \\
        action is greedy %
    };
        \node[leafnode](3;000) at (12,10) {
            \textbf{Using reward $r(s,a)$ only:} \\
            \citet[\alg{4}]{singh_1994_avgrewrl}, \\
            \citet[\equ{8}]{das_1999_smart}
        };
        \node[leafnode](3;001) at (13,6) {
            \textbf{Using reward $r(s,a)$ with adjustment:} \\
            \citet[\secc{5}]{schwartz_1993_rlearn}
        };
    \node[nonleafnode](2;01) at (8,2) {
        Update at every action %
    };
        \node[leafnode](3;010) at (13,4) {
            \textbf{Using reward $r(s,a)$ with adjustment:} \\ %
            \citet[\alg{3}]{singh_1994_avgrewrl}, \\
            \citet[\equ{6}]{wan_2020_avgrew}
        };
        \node[leafnode](3;011) at (13,0) {
            \textbf{Using a state reference $\sref$:} \\
            \citet[\page{404}]{bertsekas_1996_neurodp}, \\
            \citet[\equ{2.9b}]{abounadi_2001_rviqlearn}, \\
            \citet[\page{551}]{bertsekas_2012_dpoc}
        };
\node[nonleafnode](1;1) at (3, -4) {non-SA \\ (1-timescale $Q_b$-learning) };
    \node[leafnode](2;10) at (9,-2) {
        \textbf{Using state $\sref$ and action $\aref$ references:} \\
        \citet[\secc{2.2}]{abounadi_2001_rviqlearn}, \\
        \citet[\equ{8, 10}]{prashanth_2011_tlrl}
    };
    \node[leafnode](2;11) at (9,-6) {
        \textbf{Not using any references:} \\ %
        \citet[\secc{2.2}]{abounadi_2001_rviqlearn}, \\
				\citet{yang_2016_csv}, \\ %
        \citet[\equ{11}]{avrachenkov_2020_wiql}, \\
        \citet[\secc{3}]{jafarniajahromi_2020_nor}  %
    };

\foreach \i in {0,1}{ \draw[edge] (0;0.east) -- (1;\i.west);}
\foreach \i in {0,1}{
    \foreach \j in {0,1}{ \draw[edge] (1;\i.east) -- (2;\i\j.west);}
}
\foreach \i in {0}{
    \foreach \j in {0,1}{
        \foreach \k in {0,1}{\draw[edge] (2;\i\j.east) -- (3;\i\j\k.west);}
    }
}
\end{tikzpicture}
\caption{Taxonomy for optimal-gain approximation $\hat{v}_g^*$,
used in value-iteration based average-reward model-free RL.
Here, SA stands for stochastic approximation.
For details, see \secref{sec:valiter_tabular}.}
\label{fig:taxonomy_optgainapprox}
\end{figure}
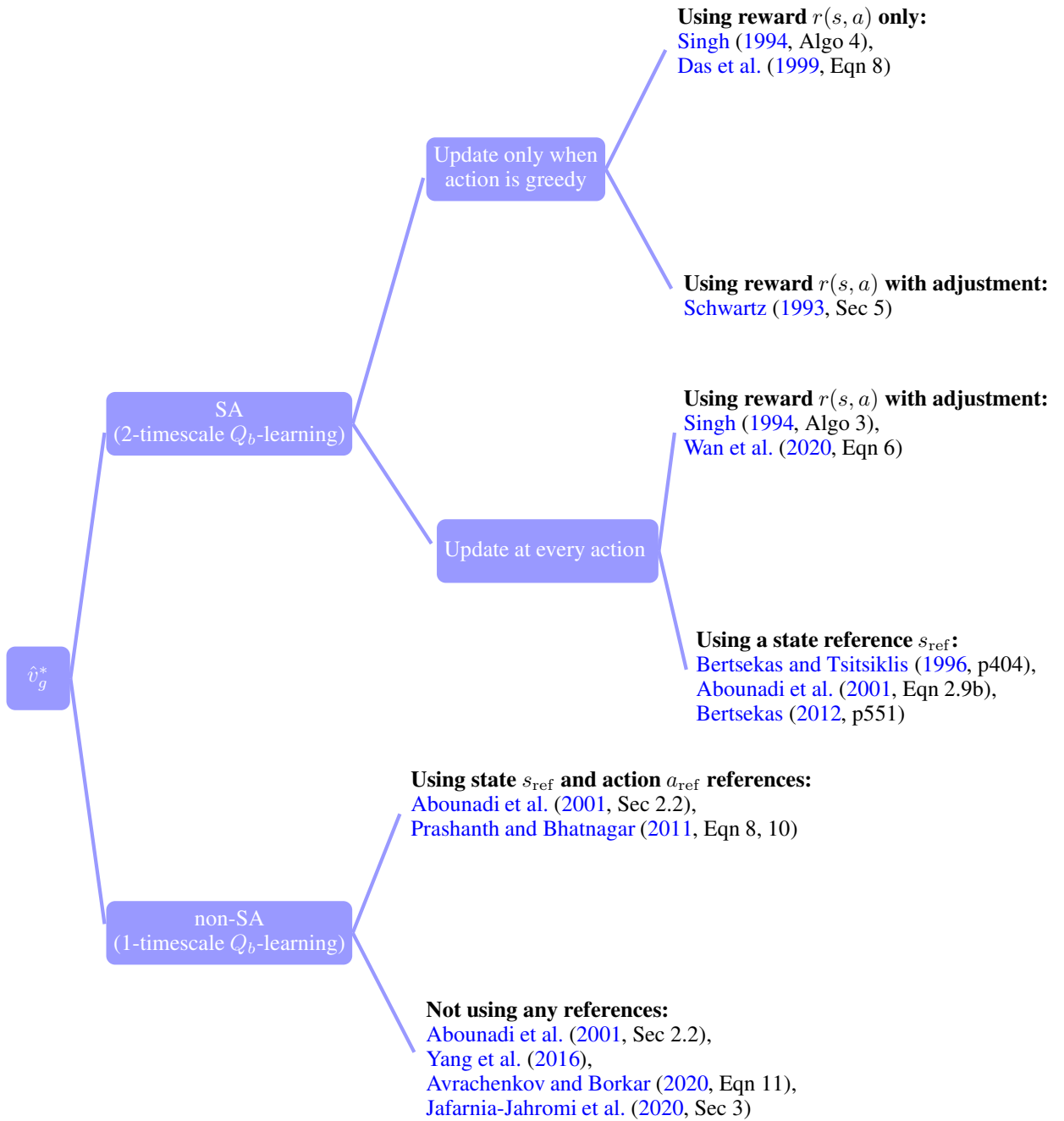
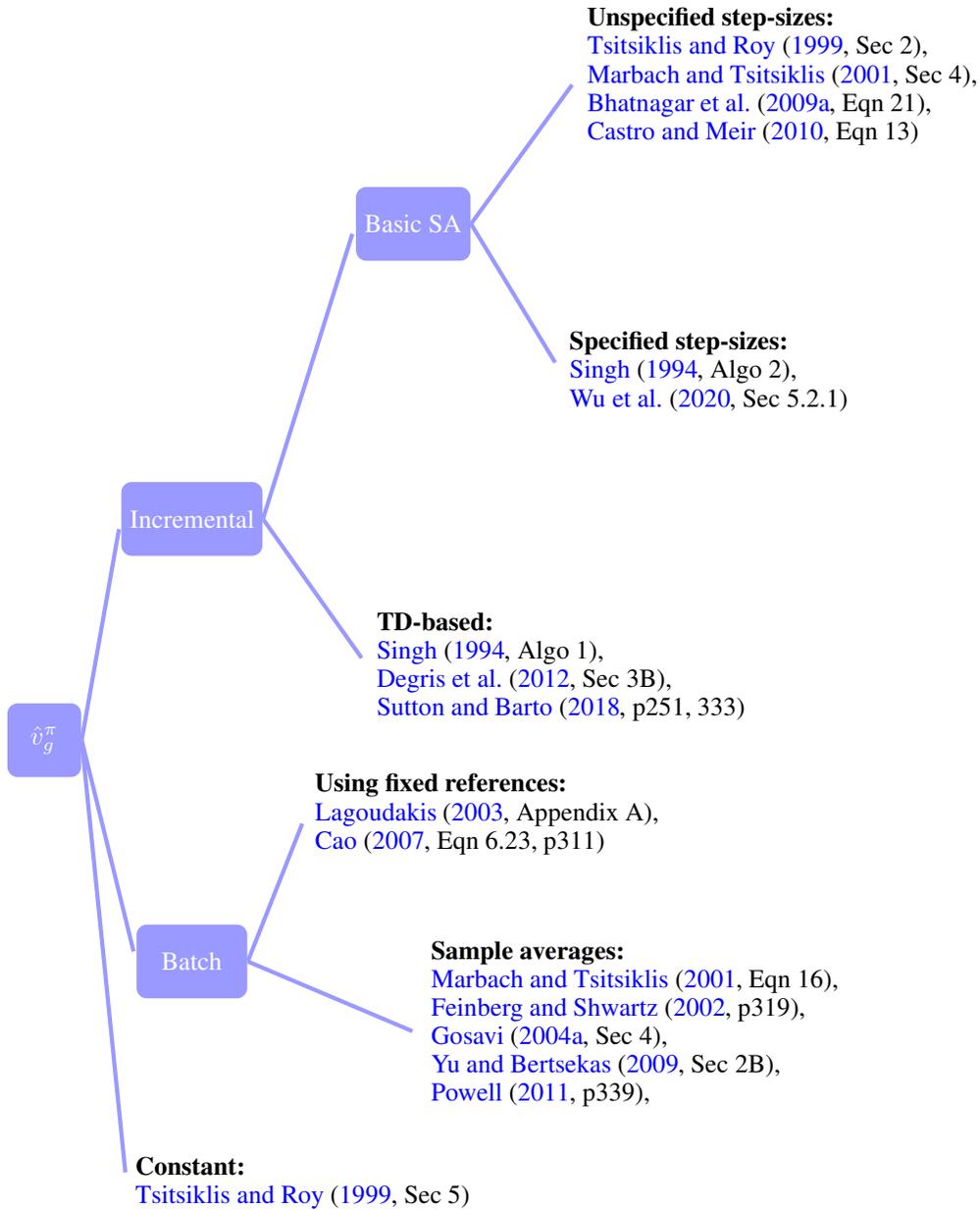
\begin{figure}
\begin{tikzpicture}[
rootnode/.style = {
    shape = rectangle, rounded corners,
    fill           = blue!40!white,
    minimum width  = 1cm,
    minimum height = 1cm,
    align          = center,
    text           = white},
nonleafnode/.style = {
    shape = rectangle, rounded corners,
    fill           = blue!40!white,
    minimum width  = 1.5cm,
    minimum height = 1cm,
    align          = center,
    text           = white},
leafnode/.style = {
    shape = rectangle,
    fill           = white,
    minimum width  = 3cm,
    minimum height = 1.5cm,
    align          = left,
    text           = black},
edge/.style  = {
    -,
    ultra thick,
    blue!40!white,
    shorten >= 4pt}
]

\node[rootnode](0;0) at (0, 0) {$\hat{v}_g^\pi$};
\node[nonleafnode](1;0) at (2,3) {Incremental};
    \node[nonleafnode](2;00) at (5,7) {Basic SA};
        \node[leafnode](3;000) at (10,9) {
            \textbf{Unspecified step-sizes:} \\
            \citet[\secc{2}]{tsitsiklis_1999_avgtd},\\
            \citet[\secc{4}]{marbach_2001_simopt},\\
            \citet[\equ{21}]{bhatnagar_2009_nac}, \\
            \citet[\equ{13}]{castro_2010_stsac}
        };
        \node[leafnode](3;001) at (9,5) {
            \textbf{Specified step-sizes:} \\
            \citet[\alg{2}]{singh_1994_avgrewrl}, \\
            \citet[\secc{5.2.1}]{wu_2020_ttsac}
        };
    \node[leafnode](2;01) at (7,1) {
        \textbf{TD-based:} \\
        \citet[\alg{1}]{singh_1994_avgrewrl}, \\
        \citet[\secc{3B}]{degris_2012_contact}, \\
        \citet[\page{251, 333}]{sutton_2018_irl}
    };
\node[nonleafnode](1;1) at (2,-3) {Batch};
    \node[leafnode](2;10) at (6,-1) {
        \textbf{Using fixed references:} \\
        \citet[\app{A}]{lagoudakis_2003_thesis}, \\
        \citet[\equ{6.23}, \page{311}]{cao_2007_slo}
    };
    \node[leafnode](2;11) at (8,-4) {
        \textbf{Sample averages:} \\
        \citet[\equ{16}]{marbach_2001_simopt}, \\
        \citet[\page{319}]{feinberg_2002_hmdp}, \\
        \citet[\secc{4}]{gosavi_2004_qplearn}, \\
        \citet[\secc{2B}]{yu_2009_lspe},\\
        \citet[\page{339}]{powell_2011_adp}, \\
    };
\node[leafnode](1;2) at (3.5,-6) {
    \textbf{Constant:} \\
    \citet[\secc{5}]{tsitsiklis_1999_avgtd}
};

\foreach \i in {0,1,2}{ \draw[edge] (0;0.east) -- (1;\i.west);}

\foreach \i in {0,1}{
    \foreach \j in {0,1}{ \draw[edge] (1;\i.east) -- (2;\i\j.west);}
}

\foreach \i in {0}{
    \foreach \j in {0}{
        \foreach \k in {0,1}{\draw[edge] (2;\i\j.east) -- (3;\i\j\k.west);}
    }
}

\end{tikzpicture}
\caption{Taxonomy for gain approximation $\hat{v}_g^\pi$,
used in policy-iteration based average-reward model-free RL.
Here, SA stands for stochastic approximation, whereas TD for temporal difference.
For details, see \secref{sec:politer_gainapprox}.}
\label{fig:taxonomy_gain}
\end{figure}
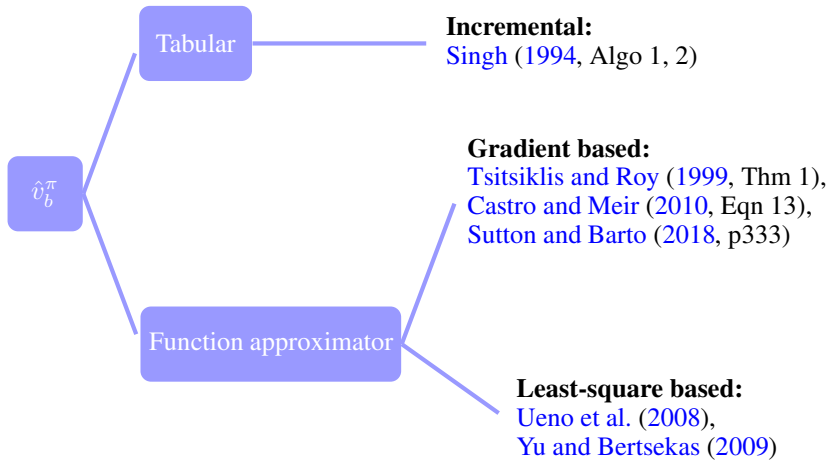
\begin{figure}
\begin{tikzpicture}[
rootnode/.style = {
    shape = rectangle, rounded corners,
    fill           = blue!40!white,
    minimum width  = 1cm,
    minimum height = 1cm,
    align          = center,
    text           = white},
nonleafnode/.style = {
    shape = rectangle, rounded corners,
    fill           = blue!40!white,
    minimum width  = 1.5cm,
    minimum height = 1cm,
    align          = center,
    text           = white},
leafnode/.style = {
    shape = rectangle,
    fill           = white,
    minimum width  = 3cm,
    minimum height = 1.5cm,
    align          = left,
    text           = black},
edge/.style  = {
    -,
    ultra thick,
    blue!40!white,
    shorten >= 4pt}
]

\node[rootnode](0;0) at (0, 0) {$\hat{v}_b^\pi$};
\node[nonleafnode](1;0) at (2,2) {Tabular};
    \node[leafnode](2;00) at (7,2) {
        \textbf{Incremental:} \\
        \citet[\alg{1, 2}]{singh_1994_avgrewrl}
    };
\node[nonleafnode](1;1) at (3,-2) {Function approximator};
    \node[leafnode](2;10) at (8,0) {
        \textbf{Gradient based:} \\
        \citet[\thm{1}]{tsitsiklis_1999_avgtd}, \\
        \citet[\equ{13}]{castro_2010_stsac}, \\
        \citet[\page{333}]{sutton_2018_irl}
    };
    \node[leafnode](2;11) at (8,-3) {
        \textbf{Least-square based:} \\
        \cite{ueno_2008_lstd},\\
        \cite{yu_2009_lspe}
    };

\foreach \i in {0,1}{ \draw[edge] (0;0.east) -- (1;\i.west);}
\foreach \i in {0,1}{
    \foreach \j in {0}{ \draw[edge] (1;\i.east) -- (2;\i\j.west);}
}
\draw[edge] (1;1.east) -- (2;11.west);

\end{tikzpicture}
\caption{Taxonomy for state-value approximation $\hat{v}_b^\pi$,
used in policy-iteration based average-reward model-free RL.
For details, see \secref{sec:politer_valueapprox}.}
\label{fig:taxonomy_stateval}
\end{figure}
\begin{figure}
\begin{tikzpicture}[
rootnode/.style = {
    shape = rectangle, rounded corners,
    fill           = blue!40!white,
    minimum width  = 1cm,
    minimum height = 1cm,
    align          = center,
    text           = white},
nonleafnode/.style = {
    shape = rectangle, rounded corners,
    fill           = blue!40!white,
    minimum width  = 1.5cm,
    minimum height = 1cm,
    align          = center,
    text           = white},
leafnode/.style = {
    shape = rectangle,
    fill           = white,
    minimum width  = 3cm,
    minimum height = 1.5cm,
    align          = left,
    text           = black},
edge/.style  = {
    -,
    ultra thick,
    blue!40!white,
    shorten >= 4pt}
]

\node[rootnode](0;0) at (0, 0) {$\hat{q}_b^\pi$};
\node[nonleafnode](1;0) at (2,4) {Tabular};
    \node[leafnode](2;00) at (6,5.5) {
        \textbf{Incremental:} \\
        \cite{sutton_2000_pgfnapprox},\\
        \cite{marbach_2001_simopt}, \\
        \cite{gosavi_2004_qplearn}
    };
    \node[leafnode](2;01) at (7,3.5) {
        \textbf{Batch:} \\
        \citet[\secc{1}]{sutton_2000_pgfnapprox},\\
        \citet[\secc{6}]{marbach_2001_simopt}, \\
        \citet[\alg{3}]{wei_2019_avgrew}
    };
\node[nonleafnode](1;1) at (2,-3) {Function \\ approximator};
    \node[nonleafnode](2;10) at (5,-1) {Gradient based};
        \node[leafnode](3;100) at (11,1) {
            \textbf{Parameterized action values $\hat{q}_b^\pi(\vecb{w})$} \\
            \textbf{(using $\vecb{\theta}$-compatible $\vecb{f}_{\vecb{\theta}}(s, a)$),
                except stated otherwise):} \\
            \citet[\thm{2}]{sutton_2000_pgfnapprox}, \\ %
            \citet[\thm{1}]{kakade_2002_npg}, \\ %
            \citet[\equ{3.1}]{konda_2003_aca},  \\ %
            \citet[\page{251}]{sutton_2018_irl} (unspecified $\vecb{f}(s, a)$) \\ %
        };
        \node[leafnode](3;101) at (11,-2) {
            \textbf{Parameterized action advantages $\hat{\adv}_b^\pi(\vecb{w})$} \\
            \textbf{(using $\vecb{\theta}$-compatible $\vecb{f}_{\vecb{\theta}}(s, a)$):} \\
            \citet[\equ{30}, \alg{3, 4}]{bhatnagar_2009_nac} \\ %
            \citet[\equ{13}]{heess_2012_acrl}, \\ %
            \citet[\equ{17}]{iwaki_2019_i2nac} \\ %
        };
        \node[leafnode](3;102) at (11,-4) {
            \textbf{TD approximates action advantages,
                \ie $\hat{\delta}_{v_b}^\pi(\vecb{w}_v) \approx \adv_b^\pi$:} \\
            \citet[\alg{1, 2}]{bhatnagar_2009_nac}, \\ %
            \citet[\page{333}]{sutton_2018_irl} %
        };
    \node[leafnode](2;11) at (6,-5.5) {
        \textbf{Least-square based:} \\
        \citet[\app{A}]{lagoudakis_2003_thesis}
    };

\foreach \i in {0,1}{ \draw[edge] (0;0.east) -- (1;\i.west);}
\foreach \i in {0,1}{
    \foreach \j in {0,1}{\draw[edge] (1;\i.east) -- (2;\i\j.west);}
}
\foreach \i in {1}{
    \foreach \j in {0}{
        \foreach \k in {0,1,2}{\draw[edge] (2;\i\j.east) -- (3;\i\j\k.west);}
    }
}

\end{tikzpicture}
\caption{Taxonomy for action-value related approximation,
\ie action values $\hat{q}_b^\pi$ and action advantages $\hat{\adv}_b^\pi$,
used in policy-iteration based average-reward model-free RL.
Here, TD stands for temporal difference.
For details, see \secref{sec:politer_valueapprox}.}
\label{fig:taxonomy_actval}
\end{figure}
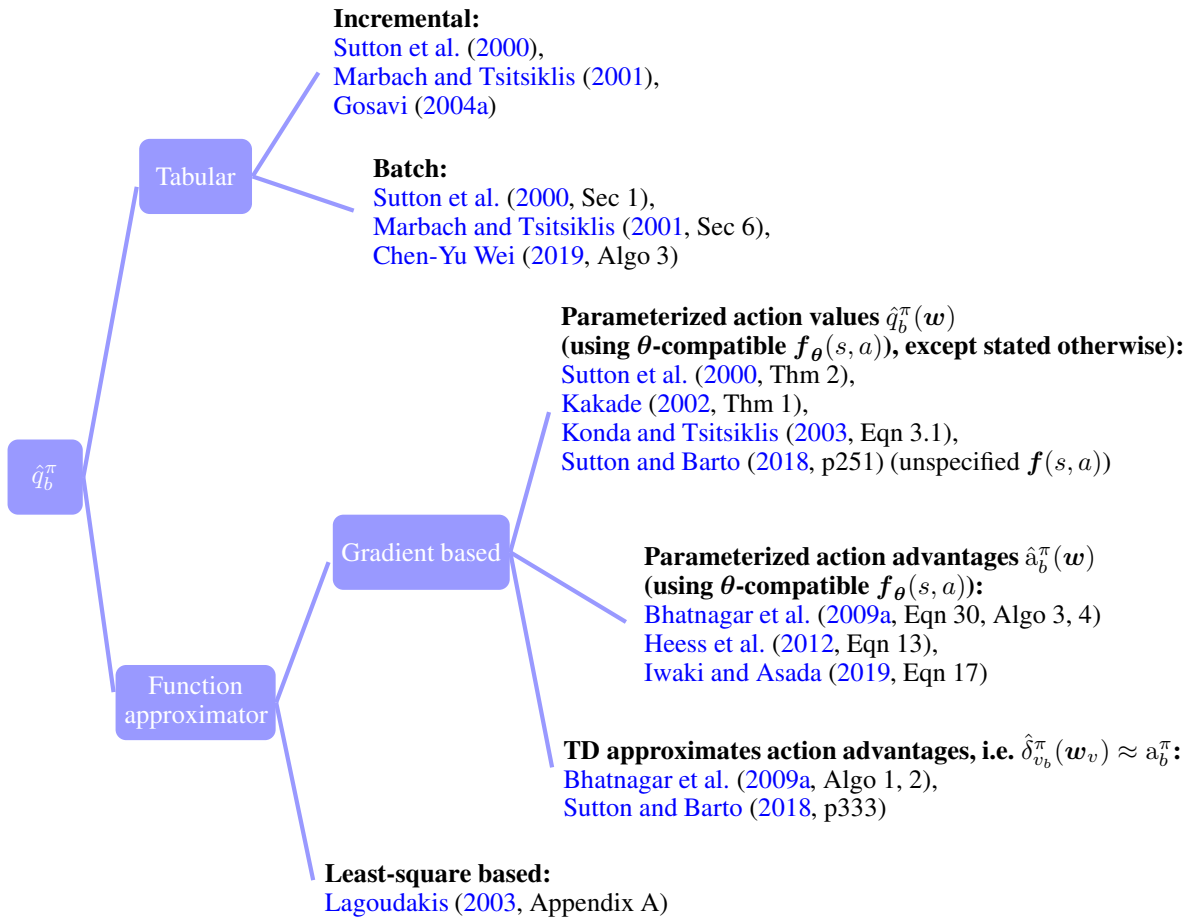
\clearpage%
}

\afterpage{%
\clearpage%

\newcounter{valiterwork_counter}
\setcounter{valiterwork_counter}{0}
\newcommand\stepvaliterworkcounter{\stepcounter{valiterwork_counter}\arabic{valiterwork_counter}}

\begin{longtable}{p{0.14\textwidth}p{0.86\textwidth}}
\caption{Existing works on value-iteration based average-reward model-free RL.
The label \textbf{T}, \textbf{F}, or \textbf{TF} (following the work number)
indicates whether the corresponding work deals with tabular, function approximation,
or both settings, respectively}
\label{tbl:valiter_work} \\

\endfirsthead

\multicolumn{2}{c}%
{{\tablename\ \thetable{} -- continued from previous page}} \\
\midrule[1pt]
\endhead

\hline \multicolumn{2}{r}{{Continued on next page}} \\
\endfoot

\bottomrule[1.25pt]
\endlastfoot

\midrule[1pt]
\textbf{Work \stepvaliterworkcounter~(T)} &
\textbf{\cite{schwartz_1993_rlearn}}: R-learning ($Q_b$-learning) \\*
Contribution &
Derive $Q_b$-learning from its discounted reward counterpart, \ie $Q_\gamma$-learning.
Update $\hat{v}_g^*$ only when the executed action is greedy,
in order to avoid skewing the approximation due to exploratory actions.
The $\hat{v}_g^*$ update involves
$\max_{a' \in \setname{A}} \hat{q}_b^*(s', a') - \max_{a' \in \setname{A}} \hat{q}_b^*(s, a')$,
which is an adjustment factor for minimizing the variance.
\\*
Experiment &
On a 50-state MDP and a 2-cycle MDP.
$Q_b$-learning outperforms $Q_\gamma$-learning in terms of
the converged average reward $\hat{v}_g^*$ and the empirical rate of learning.
\\*

\midrule[1pt]
\textbf{Work \stepvaliterworkcounter~(T)} &
\textbf{\cite{singh_1994_avgrewrl}} \\*
Contribution &
Suggest setting $\hat{q}_b^*(\sref, \aref) \gets 0$ for an arbitrary reference state-action pair,
in order to prevent $\hat{q}_b^*$ from becoming large.
Two approximation techniques for $v_g^*$.
\textbf{First}, modified version of \cite{schwartz_1993_rlearn} based on the Bellman optimality equation,
so that updates occur at every action, hence increasing sample efficiency.
\textbf{Second}, update $\hat{v}_g^*$ as the sample average of rewards received for greedy actions.
\\*
Experiment &
On 20,100-state and 5,10-action randomly-constructed MDPs.
Both proposed algorithms are shown to converge to $v_g^*$.
No comparison to \cite{schwartz_1993_rlearn}.
\\*

\midrule[1pt]
\textbf{Work \stepvaliterworkcounter~(F)} &
\textbf{\cite{bertsekas_1996_neurodp}} \\*
Contribution &
$Q_b$-learning with function approximation (\page{404}), whose parameter $\vecb{w}$
is updated via
\begin{equation*}
\vecb{w} \gets (1 - \beta) \vecb{w}
    + \beta \Big(
            r(s, a) +  \max_{a' \in \setname{A}} \hat{q}_b^*(s', a'; \vecb{w})
            - \hat{v}_g^*
        \Big)
        \nabla \hat{q}_b^*(s, a; \vecb{w}),
\end{equation*}
for some positive stepsize $\beta$.
Suggest setting $\hat{q}_b^*(\sref, \cdot) \gets 0$ for
an arbitrary-but-fixed reference state $\sref$.
The optimal gain $v_g^*$ is estimated using
\eqreffand{equ:vgstar_sa}{equ:optgain_sa_bertsekas}.
\\*
Experiment &
None
\\*

\midrule[1pt]
\textbf{Work \stepvaliterworkcounter~(F)} &
\textbf{\cite{das_1999_smart, gosavi_2002_smart, gosavi_2004_avgrew}} \\*
Contribution &
$Q_b$-learning on semi-MDPs, named Semi-Markov Average Reward Technique (SMART).
Use a feed-forward neural network to represent $\hat{q}_b^*(\vecb{w})$
with the following update (\equ{9}),
\begin{equation*}
\vecb{w} \gets \vecb{w} + \beta \{
    r(s, a) - \hat{v}_g^* + \max_{a' \in \setname{A}} \hat{q}_b^*(s', a', \vecb{w})
    - \hat{q}_b^*(s, a; \vecb{w})
\} \nabla \hat{q}_b^*(s, a; \vecb{w}).
\end{equation*}
Decay $\beta$ (and $\epsilon$-greedy exploration) slowly to~0
according to the Darken-Chang-Moody procedure (with 2 hyperparameters).
Extended to $\lambda$-SMART using the forward-view TD($\lambda$).
\\*
Experiment &
Preventative maintenance for 1- and 5-product inventory (up to $10^{11}$ states).
SMART yields the final $\hat{v}_g^*$ whose mean lies within 4\% of the exact $v_g^*$ (\tbl{2}).
It outperforms 2 heuristic baselines (\tbl{5}),
namely operational-readiness and age-replacement.
\\*

\midrule[1pt]
\textbf{Work \stepvaliterworkcounter~(T)} &
\textbf{\cite{abounadi_2001_rviqlearn, bertsekas_2012_dpoc}}: Tabular RVI and SSP $Q_b$-learning \\*
Contribution &
\textbf{First}, asymptotic convergence for RVI $Q_b$-learning based on ODE with
the usual diminishing stepsize for SA (\secc{2.2, 3}).
Stepsize-free gain approximation, namely $\hat{v}_g^* = f(\hat{q}_b^*)$, where
$f: \real{\setsize{S} \times \setsize{A}} \mapsto \real{}$ is Lipschitz,
$f(\vecb{1}_{sa}) = 1$ and $f(\vecb{x} + c \vecb{1}_{sa}) = f(\vecb{x}) + c, \forall c \in \real{}$,
$\vecb{1}_{sa}$ denotes a constant vector of all 1's in $\real{\setsize{S} \times \setsize{A}}$
(\assume{2.2}).
For example,
\begin{equation*}
f(\hat{q}_b^*) \eqdef \hat{q}_b^*(\sref, \aref), \quad
f(\hat{q}_b^*) \eqdef \max_{a} \hat{q}_b^*(\sref, a), \quad
f(\hat{q}_b^*) \eqdef \frac{1}{\setsize{S} \setsize{A}}\sum_{s, a} \hat{q}_b^*(s, a).
\end{equation*}
\textbf{Second}, SSP $Q_b$-learning based on an observation that the gain of any stationary policy
is simply the ratio of expected total reward and expected time between
2 successsive visits to the reference state (\secc{2.3}, \equ{2.9a});
also based on the contracting value iteration (\cite{bertsekas_2012_dpoc}: \secc{7.2.3}).
Asymptotic convergence based on ODE and the usual diminishing stepsize for SA (\secc{4}).
The update for $\hat{q}_b^*$ involves $\max_{a' \in \setname{A}} \hat{q}_b^*(s', a')$,
where $\hat{q}_b^*(s', a') = 0$ if $s' = \sref$ (\cite{bertsekas_2012_dpoc}: \equ{7.15}).
The optimal gain is approximated (in \equ{2.9b}) as
\begin{equation*}
\hat{v}_g^* \gets \mathbb{P}_\kappa [
    \hat{v}_g^* + \beta_g \max_{a \in \setname{A}} \hat{q}_b^*(\sref, a) ],
\end{equation*}
where $\mathbb{P}_\kappa$ denotes the projection on to an interval $[-\kappa, \kappa]$
for some constant $\kappa$ such that $v_g^* \in (-\kappa, \kappa)$.
This $\hat{v}_g^*$ should be updated at a slower rate than $\hat{q}_b^*$.
\\*
Experiment &
None
\\*

\midrule[1pt]
\textbf{Work \stepvaliterworkcounter~(F)} &
\textbf{\cite{prashanth_2011_tlrl}} \\*
Contribution &
$Q_b$-learning with function approximation (\equ{10}),
\begin{equation*}
\vecb{w} \gets \vecb{w} + \beta \{
    r(s, a) - \hat{v}_g^* + \max_{a' \in \setname{A}} \hat{q}_b^*(s', a', \vecb{w}) \},
\quad \text{with}\
\hat{v}_g^* = \max_{a'' \in \setname{A}} \hat{q}_b^*(\sref, a''; \vecb{w}).
\end{equation*}
\\*
Experiment &
On traffic-light control.
The parameter $\vecb{w}$ does not converge, it oscillates.
The proposal resulted in worse performance, compared to an actor-critic method.
\\*
\midrule[1pt]
\textbf{Work \stepvaliterworkcounter~(TF)} &
\textbf{\cite{yang_2016_csv}}: Constant shifting values (CSVs) \\*
Contribution &
Use a constant as $\hat{v}_g^*$, which is inferred from prior knowledge,
because RVI $Q_b$-learning is sensitive to the choice of $\sref$ when the state set is large.
Argue that unbounded $\hat{q}_b^*$ (when $\hat{v}_g^* < v_g^*$) are acceptable as long as the policy converges.
Hence, derive a terminating condition (as soon as the policy is deemed stable),
and prove that such a convergence could be towards the optimal policy if CSVs are
properly chosen.
\\*
Experiment &
On 4-circle MDP, RiverSwim, Tetris.
Outperform SMART, RVI $Q_b$-learning, R-learning in both tabular and
linear function approximation (with some notion of eligibility traces).
\\*

\midrule[1pt]
\textbf{Work \stepvaliterworkcounter~(T)} &
\textbf{\cite{avrachenkov_2020_wiql}} \\*
Contribution &
Tabular RVI $Q_b$-learning for the Whittle index policy (\equ{11}).
Approximate $v_g^*$ using the average of all entries of $\hat{\vecb{q}}_b^*$
(same as \cite{abounadi_2001_rviqlearn}: \secc{2.2}).
Analyse the asymptotic convergence.
\\*
Experiment &
Multi-armed restless bandits:
4-state problem with circulant dynamics, 5-state problem with restart.
The proposal is shown to converge to the exact $v_g^*$.
\\*

\midrule[1pt]
\textbf{Work \stepvaliterworkcounter~(T)} &
\textbf{\cite{wan_2020_avgrew}} \\*
Contribution &
Tabular $Q_b$-learning without any reference state-action pair.
Show empirically that such reference retards learning and causes divergence when
$(\sref, \aref)$ is infrequently visited (\eg trasient states in unichain MDPs).
Prove its asymptotic convergence under unichain MDPs, whose key is using TD
for $\hat{v}_g^*$ estimates (same as \cite{singh_1994_avgrewrl}: \alg{3}).
\\*
Experiment &
On access-control queuing.
The proposal is empirically shown to be competitive in terms of learning curves
in comparison to RVI $Q_b$-learning.
\\*

\midrule[1pt]
\textbf{Work \stepvaliterworkcounter~(T)} &
\textbf{\cite{jafarniajahromi_2020_nor}}: Exploration Enhanced Q-learning (EE-QL) \\*
Contribution &
A regret bound of $\Ot{\sqrt{\tmaxhat}}$ with tabular non-parametric policies
in weakly communicating MDPs (more general than the typically-assumed recurrent MDPs).
The key is to use a single scalar estimate $\hat{v}_g^*$ for all states,
instead of maintaining the estimate for each state.
This concentrating approximation $\hat{v}_g^*$ is assumed to be available.
For updating $\hat{q}_b^*$, use $\beta = 1/\sqrt{k}$ for the sake of analysis,
instead of the common $\beta = 1/k$.
\\*
Experiment &
On random MDPs and RiverSwim (weakly communicating variant).
Outperform MDP-OOMD and POLITEX, as well as model-based benchmarks UCRL2 and PSRL.
\\*

\end{longtable}

\clearpage%
}

\afterpage{%
\clearpage%
\newcounter{politerwork_counter}
\setcounter{politerwork_counter}{0}
\newcommand\steppoliterworkcounter{\stepcounter{politerwork_counter}\arabic{politerwork_counter}}

\begin{longtable}{p{0.14\textwidth}p{0.86\textwidth}}
\caption{Existing works on policy-iteration based average-reward model-free RL.
The label \textbf{E}, \textbf{I}, or \textbf{EI} (following the work number)
indicates whether the corresponding work is (mainly) about \emph{on-policy} policy evaluation,
policy improvement, or both, respectively.}
\label{tbl:politer_work} \\
\endfirsthead

\multicolumn{2}{c}%
{{\tablename\ \thetable{} -- continued from previous page}} \\
\midrule[1pt]
\endhead

\hline \multicolumn{2}{r}{{Continued on next page ...}} \\
\endfoot

\bottomrule[1.25pt]
\endlastfoot

\midrule[1pt]
\textbf{Work \steppoliterworkcounter~(E)} &
\textbf{\cite{singh_1994_avgrewrl}} \\*
Contribution &
Tabular policy evaluation:
$\hat{v}_b^\pi(s) \gets \{1 - \beta_v \} \hat{v}_b^\pi(s)
    + \beta_v \{ r(s, a) + \hat{v}_b^\pi(s') - \hat{v}_g^\pi \}$,
with \emph{either}
\alg{1}: $\hat{v}_g^\pi \gets (1 - \beta_g) \hat{v}_g^\pi
    + \beta_g \{ r(s, a) + \hat{v}_b^\pi(s') - \hat{v}_b^\pi(s_t) \}$,
based on SA and Bellman expectation equation, \emph{or}
\alg{2}:
$\hat{v}_g^\pi \gets \{ (t \times \hat{v}_g^\pi) + r(s, a) \}/(t+1)$
as the sample average of the rewards received for greedy actions.
It is noted that $\beta_v$ has dependency on both timesteps and current states,
and $\beta_g$ on timesteps, but provide no further explanation.
\\*
Experiment &
On 20-, 100-state and 5-, 10-action random MDPs.
\alg{2} is better than \alg{1} with respect to the absolute total errors
relative to the exact $v_g^\pi$ (over 10 trials).
\\*

\midrule[1pt]
\textbf{Work \steppoliterworkcounter~(E)} &
\textbf{\cite{tsitsiklis_1999_avgtd}} \\*
Contribution &
On-policy TD($\lambda$) with $\hat{v}_b^\pi(\vecb{w})$ as
a linearly independent combination of fixed basis functions.
A proof of asymptotic convergence with probability~1, based on ODE and SA (\thm{1}).
A bound on approximation errors and its dependence on a mixing factor (\thm{3}).
The aforementioned convergence and error analysis are for adaptive gain estimates;
their counterpart for fixed gain estimates is also established in \thm{4}
with an additional mixing factor (for this case $\lambda$ far from 1 is preferable).
\\*
Experiment &
None
\\*

\midrule[1pt]
\textbf{Work \steppoliterworkcounter~(I)} &
\textbf{\cite{sutton_2000_pgfnapprox}}: (Randomized) Policy Gradient Theorem \\*
Contribution &
Policy gradient with function approximation for
a parameterized policy $\pi(\vecb{\theta})$, namely
$\nabla v_g(\vecb{\theta}) = \sum_{s \in \setname{S}} p^\pi(s)
\sum_{a \in \setname{A}} \hat{q}_b^\pi(s, a; \vecb{w}) \nabla \pi(s, a; \vecb{\theta})$,
with $\hat{q}_b^{\vecb{\theta}}(s, a; \vecb{w}) = \vecb{w}^\intercal \nabla \log \pi(a|s; \vecb{\theta})$,
which is said to be compatible with policy parameterization.
\thm{3} assures asymptotic convergence to a \emph{locally} optimal policy,
assuming bounded rewards, bounded $\nabla v_g(\vecb{\theta})$, and
step sizes satisfying standard SA requirements.
\\*
Experiment &
None
\\*

\midrule[1pt]
\textbf{Work \steppoliterworkcounter~(I)} &
\textbf{\cite{marbach_2001_simopt}}
\\*
Contribution &
The same randomized policy gradient (\secc{6}) as \cite{sutton_2000_pgfnapprox},
but for a regenerative process with one recurrent state $\sref$ under every deterministic policy.
Gradient updates are either every step (incremental) or every regenerative cycle (batch).
The former uses eligibility traces (\secc{5}) with $\lambda_{\vecb{\theta}} < 1$,
speeding up convergence by an order of magnitude, while introducing a negligible bias.
Using tabular $q_b^{\vecb{\theta}}$.
The estimate $\hat{v}_g^\pi$ is computed as a weighted average of all past rewards
from all regenerative cycles; resulting in lower variance (\equ{16}).
\\*
Experiment &
Call admission control, whose exact $v_g^* = 0.8868$.
Exact gradient computation yields $\hat{v}_g^* = 0.8808$ (100 updates), whereas
those with every step updates yield $\hat{v}_g^* = 0.8789$
for $\lambda_{\vecb{\theta}} = 1$ in 8 million updates, and
$\hat{v}_g^* = 0.8785$ for $\lambda_{\vecb{\theta}} = 0.99$ in 1 million updates.
\\*

\midrule[1pt]
\textbf{Work \steppoliterworkcounter~(I)} &
\textbf{\cite{kakade_2002_npg}}: Natural Policy Gradient \\*
Contribution &
Propose one possible Riemmanian-metric matrix, namely
\begin{equation*}
\mat{F}(\vecb{\theta}) \eqdef \E{S \sim p_{\vecb{\theta}}^\star }{\mat{F}(S; \vecb{\theta})},
\quad \text{with}\
\mat{F}(s; \vecb{\theta}) \eqdef \E{A}{\nabla \pi(A|s; \vecb{\theta})
\nabla^\intercal \pi(A|s; \vecb{\theta})},
\end{equation*}
which is the Fisher Information matrix of a distribution
$\pi(A| s; \vecb{\theta}), \forall s \in \setname{S}$.
Recall that there exists a probability manifold (surface) corresponding to each state $s$,
where $\pi(A| s; \vecb{\theta})$ is a point on that manifold with coordinates $\vecb{\theta}$.
This $\mat{F}(s; \vecb{\theta})$ defines the \emph{same} distance between 2 points
(on the manifold of $s$) regardless of policy parameterization (the choice of coordinates).
Prove that $\vecb{w}^* = \mat{F}(\vecb{\theta}) \nabla v_g(\vecb{\theta})$,
where $\vecb{w}^*$ minimizes the MSE and
$\hat{\vecb{q}}_b^{\vecb{\theta}}(\vecb{w})$ is $\vecb{\theta}$-compatible (\thm{1}).
For derivation of $\mat{F}(\vecb{\theta})$ via the trajectory distribution manifold,
see \cite{bagnell_2003_covps}: \thm{1}, and \cite{peters_2008_nac}: \app{A}.
\\*
Experiment &
On 2-state MDP, 1-dimensional linear quadratic regulator, and scaled-down Tetris.
Natural gradients lead to a faster rate of learning, compared to standard gradients.
\\*

\midrule[1pt]
\textbf{Work \steppoliterworkcounter~(EI)} &
\textbf{\cite{konda_2003_aca}}: A class of actor-critic algorithms \\*
Contribution &
Interpret randomized policy gradients as an inner product of
2 real-valued functions on $\setname{S} \times \setname{A}$, \ie
\begin{equation*}
\partial v_g(\vecb{\theta})/ \partial \theta_i
= \langle q_b(\vecb{\theta}), \nabla_i \log \pi(\vecb{\theta}) \rangle_{\vecb{\theta}}
= \sum_{s \in \setname{S}} \sum_{a \in \setname{A}} p_{\vecb{\theta}}^\star(s, a)
    q_b^{\vecb{\theta}}(s, a) \nabla_i \log \pi(s, a; \vecb{\theta}),
\end{equation*}
where $\nabla_i \log \pi(\vecb{\theta})$ denotes the $i$-th unit vector component
of $\nabla \log \pi(\vecb{\theta})$ for $i = 1, 2, \ldots, d_{\vecb{\theta}}$.
Thus, in order to compute $\nabla v_g(\vecb{\theta})$, it suffices to learn
the projection of $q_b(\vecb{\theta})$ onto the span of the unit vectors
$\{\nabla_i \log \pi(\vecb{\theta}); 1 \le i \le d_{\vecb{\theta}} \}$
in $\real{\setsize{S} \setsize{A}}$.
This span should be contained in the span of the basis vectors of
the linearly parameterized critic, \eg setting $\nabla_i \log \pi(\vecb{\theta})$
as such basis vectors.
Using TD($\lambda$) with backward-view eligibility traces.
The convergence analysis is based on the martingale approach.
\\*
Experiment &
None
\\*

\midrule[1pt]
\textbf{Work \steppoliterworkcounter~(E)} &
\textbf{\cite{lagoudakis_2003_thesis}}: Least-Squares TD of Q for average reward (LSTDQ-AR) \\*
Contribution &
\emph{Batch} policy evaluation as part of least-squares policy iteration.
Remark that because there is no exponential drop due to a discount factor,
the value function approximation (for average rewards) is more amenable
to fitting with linear architectures (\app{A}).
\\*
Experiment &
None
\\*

\midrule[1pt]
\textbf{Work \steppoliterworkcounter~(E)} &
\textbf{\cite{gosavi_2004_qplearn}}: QP-Learning \\*
Contribution &
Tabular GPI-based methods with implicit policy representation
(taking a greedy action with respect to relative action values).
Use TD for learning $\hat{q}_b^\pi$.
Decay the learning rate, as the inverse of the number of updates (timesteps).
Convergence analysis to optimal solution via ODE methods (\secc{6}).
This method is similar to Sarsa (\cite{sutton_2018_irl}: \page{251}),
but with batch gain approximation;
note that the action value approximation has its own disjoint batch and
is updated incrementally per timestep.
\\*
Experiment &
Airline yield management: 6 distinct cases, 10 runs (1 million timestep each).
Improvements over the commonly used heuristic (expected marginal seat revenue): 3.2\% to 16\%;
whereas over $Q_{\gamma=0.99}$-learning: 2.3\% to 9.3\%.
\\*

\midrule[1pt]
\textbf{Work \steppoliterworkcounter~(E)} &
\textbf{\cite{ueno_2008_lstd}}: gLSTD and LSTDc \\*
Contribution &
LSTD-based \emph{batch} policy evaluation (on state values) via semiparametric statistical inference.
Using the so-called estimating-function methods, leading to gLSTD and LSTDc.
Analyze asymptotic variance of linear function approximation.
\\*
Experiment &
On a 4-state, 2-action MDP.
The gLSTD and LSTDc are shown to have lower variance.
\\*

\midrule[1pt]
\textbf{Work \steppoliterworkcounter~(EI)} &
\textbf{\cite{bhatnagar_2009_nac, prashanth_2011_tlrl}}: Natural actor critic \\*
Contribution &
An iterative procedure to estimate the inverse of the Fisher information matrix $\mat{F}^{-1}$.
It is based on weighted average in SA and Sherman-Morrison formula (\equ{26}, \alg{2}).
This $\mat{F}^{-1}$ is also used as a preconditioning matrix for the critic's gradients
(\equ{35}, \alg{4}).
Use a $\vecb{\theta}$-compatible parameterized action advantage approximator (critic),
whose parameter $\vecb{w}$ is utilized as natural gradient estimates (\equ{30}, \alg{3}).
Based on ODE, prove the asymptotic convergence to a small neighborhood of
the set of local maxima of $v_g^*$.
Related to \citep{bhatnagar_2008_inac, abdulla_2007_avgrewac}.
\\*
Experiment &
On Generic Average Reward Non-stationary Environment Testbed (GARNET).
\alg{3} with parametric action advantage approximator yields
reliably good performance in both small and large GARNETs, and
outperforms that of \cite{konda_2003_aca}.
\\*

\midrule[1pt]
\textbf{Work \steppoliterworkcounter~(E)} &
\textbf{\cite{yu_2009_lspe}} \\*
Contribution &
Analyze convergence and the rate of convergence of
TD-based least squares batch policy evaluation for $v_b^\pi$,
specifically LSPE($\lambda$) algorithm for any $\lambda \in (0, 1)$
and any constant stepsize $\beta \in (0, 1]$.
The non-expansive mapping is turned to contraction through
the choice of basis functions and a constant stepsize.
\\*
Experiment &
On 2- and 100-state randomly constructed MDPs.
LSPE($\lambda$) is competitive with LSTD($\lambda$).
\\*

\midrule[1pt]
\textbf{Work \steppoliterworkcounter~(I)} &
\textbf{\cite{morimura_2009_gnac, morimura_2008_nnpg}}: generalized Natural Actor-Critic (gNAC) \\*
Contribution &
A variant of natural actor critic that uses the generalized natural gradient (gNG), \ie
\begin{equation*}
\mat{F}_{sa}(\vecb{\theta}, \kappa) \eqdef \kappa \mat{F}_s(\vecb{\theta}) + \mat{F}_a(\vecb{\theta})
\quad \text{with some constant}\ \kappa \in [0, 1].
\end{equation*}
This linear interpolation controlled by $\kappa$ can be interpreted as a continuous interpolation
with respect to the $k$-timesteps state-action joint distribution.
In particular, $\kappa = 0$ corresponds to $\infty$-timesteps
(after converging to the stationary state distribution),
whereas $\kappa = 1$ corresponds to $1$-timestep.
Provide an efficient implementation for the gNG learning based on the estimating function theory,
equipped with an auxilary function for variance reduction (\lmm{1}, \thm{1}).
The policy parameter is updated by a gNG estimate, which is a solution of the estimating function.
\\*
Experiment &
On randomly constructed MDPs with 2 actions and 2, 5, 10, up to 100 states;
using the technique of \cite{morimura_2010_sdpg} for $\nabla p_s(\vecb{\theta})$.
The generalized variant with $\kappa = 1$ converges to the same point as that
of $\kappa = 0.25$, but with a slower rate of learning.
Both outperform the baseline NAC algorithm, which is equivalent to using $\kappa = 0$.
\\*

\midrule[1pt]
\textbf{Work \steppoliterworkcounter~(EI)} &
\textbf{\cite{castro_2010_stsac, castro_2010_abase, castro_2009_tdac}} \\*
Contribution &
ODE-based convergence analysis of 1-timescale actor-critic methods to a \emph{neighborhood} of
a local maximum of $v_g^*$ (instead of to a local maximum itself as in 2-timescale variants).
This single timescale is motivated by a biological context, \ie
unclear justification for 2 timescales operating within the same anatomical structure.
\\*
Experiment &
On GARNET, standard gradients, and linearly parameterized TD($\lambda$) critic with eligibility trace.
During the initial phase, 1-timescale algorithm converges faster than
that of \cite{bhatnagar_2009_nac}, whereas the long term behavior is problem-dependent.
\\*

\midrule[1pt]
\textbf{Work \steppoliterworkcounter~(I)} &
\textbf{\cite{matsubara_2010_lrpg}}: Adaptive stepsizes \\*
Contribution &
An adaptive stepsize for both standard and natural policy gradient methods.
It is achieved by setting the distance between $\vecb{\theta}$ and
$(\vecb{\theta} + \Delta \vecb{\theta})$ to some \emph{user-specified} constant,
where the effect of a change $\Delta \vecb{\theta}$ on $v_g$ is measured by a Riemannian metric.
Thus, $\alpha(\vecb{\theta}) =
1 / \{\nabla^\intercal v_g(\vecb{\theta}) \mat{F}^{-1}(\vecb{\theta}) \nabla v_g(\vecb{\theta})\}$.
Estimate $\mat{F}(\vecb{\theta})$ as the mean of the exponential recency-weighted average,
\ie $\hat{\mat{F}} \gets \hat{\mat{F}}
+ \lambda (\nabla v_g(\vecb{\theta}) \nabla^\intercal v_g(\vecb{\theta})  - \hat{\mat{F}})$.
\\*
Experiment &
On 3- and 20-state MDPs, with eligibility trace on policy gradients.
The proposed adaptive natural gradients lead to faster convergence than
that of \cite{kakade_2002_npg}.
\\*

\midrule[1pt]
\textbf{Work \steppoliterworkcounter~(I)} &
\textbf{\cite{heess_2012_acrl}}: Energy-based policy parameterization \\*
Contribution &
Energy-based policy parameterization with latent variables (\equ{3}), which enables
\emph{complex} non-linear relationship between actions and states.
Incremental approximation techniques for the partition function and
for two expectations used in computing compatible features (\equ{19}).
\\*
Experiment &
Octopus arm (continuous states, discrete actions), compatible action advantage approximator.
The proposal (which builds upon \cite{bhatnagar_2009_nac})
ourperforms the non-linear version of the Sarsa algorithm (\equ{5}).
\\*

\midrule[1pt]
\textbf{Work \steppoliterworkcounter~(EI)} &
\textbf{\cite{degris_2012_contact}} \\*
Contribution &
Empirical study of actor-critic methods with backward-view eligibility traces
for both actor and critic.
Parameterize continuous action policies using a normal distribution,
whose variance is used to scale the gradient.
\\*
Experiment &
On swing-up pendulum (continuous states and actions).
Use standard and natural gradients, and linearly parameterized state-value approximator
with tile coding for state feature extractions.
The use of variance-scaled gradients and eligibility traces
significantly improves performance, but the use of natural gradients does not.
\\*

\midrule[1pt]
\textbf{Work \steppoliterworkcounter~(I)} &
\textbf{\cite{morimura_2014_pgrl}} \\*
Contribution &
Regularization for policy gradient updates, \ie $- \kappa \nabla h(\vecb{\theta})$,
where $\kappa$ denote some scaling factor, and $h$ the hitting time, which
controls the magnitude of the mixing time.
A TD-based approximation for the gradient of the hitting time, $\nabla h(\vecb{\theta})$.
\\*
Experiment &
On a 2-state MDP.
The regularized policy gradient leads to faster learning,
compared to non-regularized ones (both standard and natural gradients).
It also reduces the effect of policy initialization on the learning performance.
\\*

\midrule[1pt]
\textbf{Work \steppoliterworkcounter~(I)} &
\textbf{\cite{furmston_2016_newtonps, furmston_2012_uppg}} \\*
Contribution &
Estimate the Hessian of $v_g(\vecb{\theta})$ via Gauss-Newton methods (\alg{2} in \app{B.1}).
Use eligibility traces for both the gradients and the Hessian approximates.
\\*
Experiment &
None
\\*

\midrule[1pt]
\textbf{Work \steppoliterworkcounter~(EI)} &
\textbf{\cite{sutton_2018_irl}} \\*
Contribution &
Present 2 algorithms:
a) semi-gradient Sarsa with $\hat{q}_b^\pi(\vecb{w})$ (\page{251}), and
b) actor-Critic with per-step updates and backward-view eligibility traces
for both actor and critic $\hat{v}_b^\pi(\vecb{w})$ (\page{333}).
\\*
Experiment &
Access-control queueing with tabular action value approximators (\page{252}).
\\*

\midrule[1pt]
\textbf{Work \steppoliterworkcounter~(E)} &
\textbf{\cite{devra_2018_dtdl, devraj_2016_dtfvf}}: grad-LSTD($\lambda$) \\*
Contribution &
Grad-LSTD($\lambda$) for
$\vecb{w}^* = \argmin_{\vecb{w}} \| \hat{v}_b^\pi(\vecb{w}) - v_b^\pi \|_{p_s^\pi, 1}$,
which is a quadratic program (\equ{30, 37}), instead of
$\vecb{w}^* = \argmin_{\vecb{w}} \| \hat{v}_b^\pi(\vecb{w}) - v_b^\pi \|_{p_s^\pi}^2$
in the standard LSTD(1). %
Analysis of convergence and error rate for linear parameterization.
Claim that grad-LSTD($\lambda$) is applicable for models that do \emph{not} have regeneration.
\\*
Experiment &
On a single-server queue with controllable service rate.
With respect to the Bellman error, grad-LSTD($\lambda$) appears to converge faster
than LSTD($\lambda$).
Moreover, the variance of grad-LSTD's parameters is smaller compared to those of LSTD.
\\*

\midrule[1pt]
\textbf{Work \steppoliterworkcounter~(EI)} &
\textbf{\cite{abbasi-yadkori_2019_politex, abbasi-yadkori_2019_xpert}}:
    POLITEX (Policy Iteration with Expert Advice) \\*
Contribution &
Reduction of controls to expert predictions:
in each state, there exists an expert algorithm and the policy's value losses are
fed to the learning algorithm.
Each policy is represented as a Boltzmann distribution over the sum of
action-value function estimates of \emph{all} previous policies.
That is, $\pi^{i + 1}(a|s) \propto \exp(- \kappa \sum_{j = 1}^i \hat{q}_b^j(s, a))$,
where $\kappa$ is a positive learning rate and $i$ indexes the policy iteration.
Achieve a regret bound of $\Ot{\tmax^{3/4}}$, which scales only in the number of features.
\\*
Experiment
&
On 4- and 8-server queueing with linear action value approximation (Bertsekas's LSPE):
POLITEX achieves similar performance to LSPI, and slightly outperforms RLSVI.
On Atari Pacman using non-linear TD-based action-value approximation (3 random seeds):
POLITEX obtains higher scores than DQN, arguably due to more stable learning.
\\*

\midrule[1pt]
\textbf{Work \steppoliterworkcounter~(E)} &
\textbf{\cite{iwaki_2019_i2nac}}: Implicit incremental natural actor critic (I2NAC) \\*
Contribution &
A preconditioning matrix for the gradient used in $\hat{\adv}_b^\pi(\vecb{w})$ updates,
namely: \newline
$\mat{I} - \kappa (\vecb{f}(s, a) \vecb{f}^\intercal(s, a))/(1 + \kappa \| \vecb{f}(s, a) \|^2)$,
for some scaling $\kappa \ge \beta$ and some possible use of eligibility traces
on $\vecb{f}$ (\equ{27}).
It is derived from the so-called implicit TD method with linear function approximation.
Provide asymptotic convergence analysis and show that the proposal is
less sensitive to the learning rate and state-action features.
\\*
Experiment &
None (the experiment is all in discounted reward settings)
\\*

\midrule[1pt]
\textbf{Work \steppoliterworkcounter~(EI)} &
\textbf{\cite{qiu_2019_ftac}} \\*
Contribution &
A non-asymptotic convergence analysis of actor-critic algorithm with
linearly-parameterized TD-based state-value approximation.
The keys are to interpret the actor-critic algorithm as
a bilevel optimization problem (\prop{3.2}), namely
\begin{equation*}
\max_{\vecb{\theta} \in \Theta} v_g(\pi(\vecb{\theta})),
\quad \text{subject to}
\argmin_{\vecb{w} \in \setname{W}, v_g \in \real{}} \ell(\vecb{w}, v_g; \vecb{\theta}),
\text{for some loss function $\ell$},
\end{equation*}
and to decouple the actor (upper-level) and critic (lower-level) updates;
instead of typical ODE-based techniques.
Here, ``decoupling'' implies that at every iteration, the critic \emph{start from scratch}
to estimate the value of the actor's policy.
Actor converges sublinearly in expectation to a stationary point,
although its updates are with biased gradient estimates (due to critic approximation).
The analysis assumes that the actor receives i.i.d samples.
\\*
Experiment &
None
\\*

\midrule[1pt]
\textbf{Work \steppoliterworkcounter~(EI)} &
\textbf{\cite{wei_2019_avgrew}}: Optimistic online mirror descent (MDP-OOMD) \\*
Contribution &
A regret bound of $\Ot{\sqrt{\tmax}}$ with
non-parametric tabular randomized policies in weakly communicating MDPs.
The key idea is to maintain an instance of an adversarial multi-armed bandit
at each state to learn the best action.
Assume that $\tmax$ is large enough so that mixing and hitting times are
both smaller than $\tmax/4$.
\\*
Experiment &
On a random MDP and JumpRiverSwim (both have 6 states and 2 actions).
MDP-OOMD outperforms standard Q-learning, but is on par with POLITEX.
\\*

\midrule[1pt]
\textbf{Work \steppoliterworkcounter~(EI)} &
\textbf{\cite{wu_2020_ttsac}} \\*
Contribution &
A non-asymptotic analysis of 2-timescale actor-critic methods
under non-i.i.d Markovian samples and with linearly-parameterized critic $\hat{v}_b^\pi(\vecb{w})$.
Show convergence guaraantee to an $\epsilon$-approximate 1st order stationary point
of the non-concave policy value function,
\ie $\| \nabla v_g(\vecb{\theta}) \|_2^2 \le \epsilon$,
in at most $\Ot{\epsilon^{-2.5}}$ samples when per iteration sample is 1 (\cor{4.10}).
Establish that for gain approximation with $\beta_t = 1/(1+t)^{\kappa_1}$, we have
\begin{equation*}
\sum_{t=\tau}^{\tmaxhat} \E{}{(\hat{v}_g^{\vecb{\theta}} - v_g^{\vecb{\theta}})^2}
=\bigO{\tmaxhat^{\kappa_1}} + \bigO{\tmaxhat^{1 - \kappa_1} \log \tmaxhat }
+ \bigO{\tmaxhat^{1 - 2(\kappa_2 - \kappa_1)}},
\end{equation*}
for some positive constants  $0 < \kappa_1 < \kappa_2 < 1$ that indicate
the relationship between actor's and critic's update rates (\secc{5.2.1}).
\\*
Experiment &
None
\\*

\midrule[1pt]
\textbf{Work \steppoliterworkcounter~(EI)} &
\textbf{\cite{hao_2020_aapi}}: Adaptive approximate policy iteration (AAPI) \\*
Contribution &
A learning scheme that operates in batches (phases) and achieves $\Ot{\tmax^{2/3}}$ regret bound.
It uses adaptive, data- and state-dependent learning rates, as well as
side-information (\equ{4.1}), \ie a vector computable based on past information
and being predictive of the next loss (here, action values).
The policy improvement is based on the adaptive optimistic follow-the-regularized-leader (AO-FTRL),
which can be interpreted as regularizing each policy by the KL-divergence to the previous policy.
\\*
Experiment &
On randomly-constructed MDPs with 5, 10, 20 states, and 5, 10, 20 actions;
DeepSea environment with grid-based states and 2 actions;
CartPole with continuous states and 2 actions.
AAPI outperforms POLITEX and RLSVI.
Remark that smaller phase lengths perform better, whereas
adaptive per-state learning rate is not effective in CartPole.
\\*

\end{longtable} %

\clearpage%
}

\clearpage%
}

\end{document}